\documentclass[]{wam}
\usepackage[utf8]{inputenc}
\usepackage[T1]{fontenc}
\usepackage{lmodern}
\usepackage{textcomp}
\usepackage{amsmath,amssymb,amsfonts,amsthm}
\usepackage{booktabs,multirow,array,tabularx}
\usepackage{adjustbox}
\usepackage[table]{xcolor}
\graphicspath{{figures/}{supplementary_figures/}}
\newcommand{\clip}{\operatorname{clip}}

\newcommand{\R}{\mathbb{R}}
\newcommand{\suppsec}[1]{Appendix~\ref{#1}}
\renewcommand\affiliationformat[2][]{\par{\normalsize $^{#1}$#2}}
\providecommand{\IEEEPARstart}[2]{#1#2}
\providecommand{\keywords}[1]{\noindent\textbf{Keywords:} #1}

\newcolumntype{Y}{>{\raggedright\arraybackslash}X}
\makeatletter

\newenvironment{inlinefigure}{\par\medskip\noindent\begin{minipage}{\columnwidth}\def\@captype{figure}}{\end{minipage}\par\medskip}
\makeatother
\hypersetup{pdfauthor={Dongli Wu, Jingyu Hu, Runsong Zhu, Weiming Wang, Xiaobao Wei, Chengwen Luo, Jianqiang Li, Wufan Zhao, Zhengzhe Liu},pdftitle={PhyMix: Towards Physically Consistent Single-Image 3D Indoor Scene Generation with Implicit-Explicit Optimization}}
\title{PhyMix: Towards Physically Consistent Single-Image 3D Indoor Scene Generation with Implicit--Explicit Optimization}
\author[1,2]{Dongli Wu}
\author[3]{Jingyu Hu}
\author[3]{Runsong Zhu}
\author[4]{Weiming Wang}
\author[5]{Xiaobao Wei}
\author[6]{Chengwen Luo}
\author[6]{Jianqiang Li}
\author[2]{Wufan Zhao}
\author[1,*]{Zhengzhe Liu}
\affiliation[1]{Lingnan University, Hong Kong, China}
\affiliation[2]{The Hong Kong University of Science and Technology (Guangzhou), Guangzhou, China}
\affiliation[3]{The Chinese University of Hong Kong, Hong Kong, China}
\affiliation[4]{Hong Kong Metropolitan University, Hong Kong, China}
\affiliation[5]{Peking University, Beijing, China}
\affiliation[6]{Shenzhen University, Shenzhen, China}
\contribution[*]{Corresponding author}
\correspondence{Zhengzhe Liu, \email{zhengzheliu@ln.edu.hk}}
\abstract{Existing methods for single-image 3D indoor scene generation often produce visually plausible results that violate real-world physical constraints, limiting their reliability in robotics, embodied AI, and virtual environment design.
To bridge this gap, we introduce a Physics Evaluator that assesses four aspects---geometric priors, contact, stability, and deployability---through nine sub-constraints, establishing a benchmark for evaluating physical consistency.
Our analysis reveals that even state-of-the-art methods produce scenes with substantial physical inconsistencies despite their visual plausibility.
To address these limitations, we propose a framework that incorporates feedback from the Physics Evaluator into both training and inference to improve the physical plausibility of generated scenes.
Specifically, \textbf{PhyMix} comprises two components: (i) \emph{implicit alignment} via Scene-GRPO, a critic-free group-relative preference optimization strategy that uses physical evaluations to guide the generator toward physically plausible layouts, and (ii) \emph{explicit refinement} via verifier-guided agentic repair, which translates evaluator diagnostics into executable scene-editing skills to correct residual violations at inference time.  
PhyMix unifies physical evaluation, preference-based alignment, and agentic inference-time correction to improve physical plausibility while retaining the observed scene content.
Experiments show improvements in reconstruction quality and physical plausibility, while qualitative examples on stylized and real-world images illustrate the applicability of the framework beyond synthetic inputs.}
\begin{document}
\maketitle
\keywords{Scene generation, physical plausibility, preference optimization, agentic repair.}

\section{Introduction}
\label{INTRO}

\begin{figure*}[!t]
\centering
    \includegraphics[width=\linewidth]{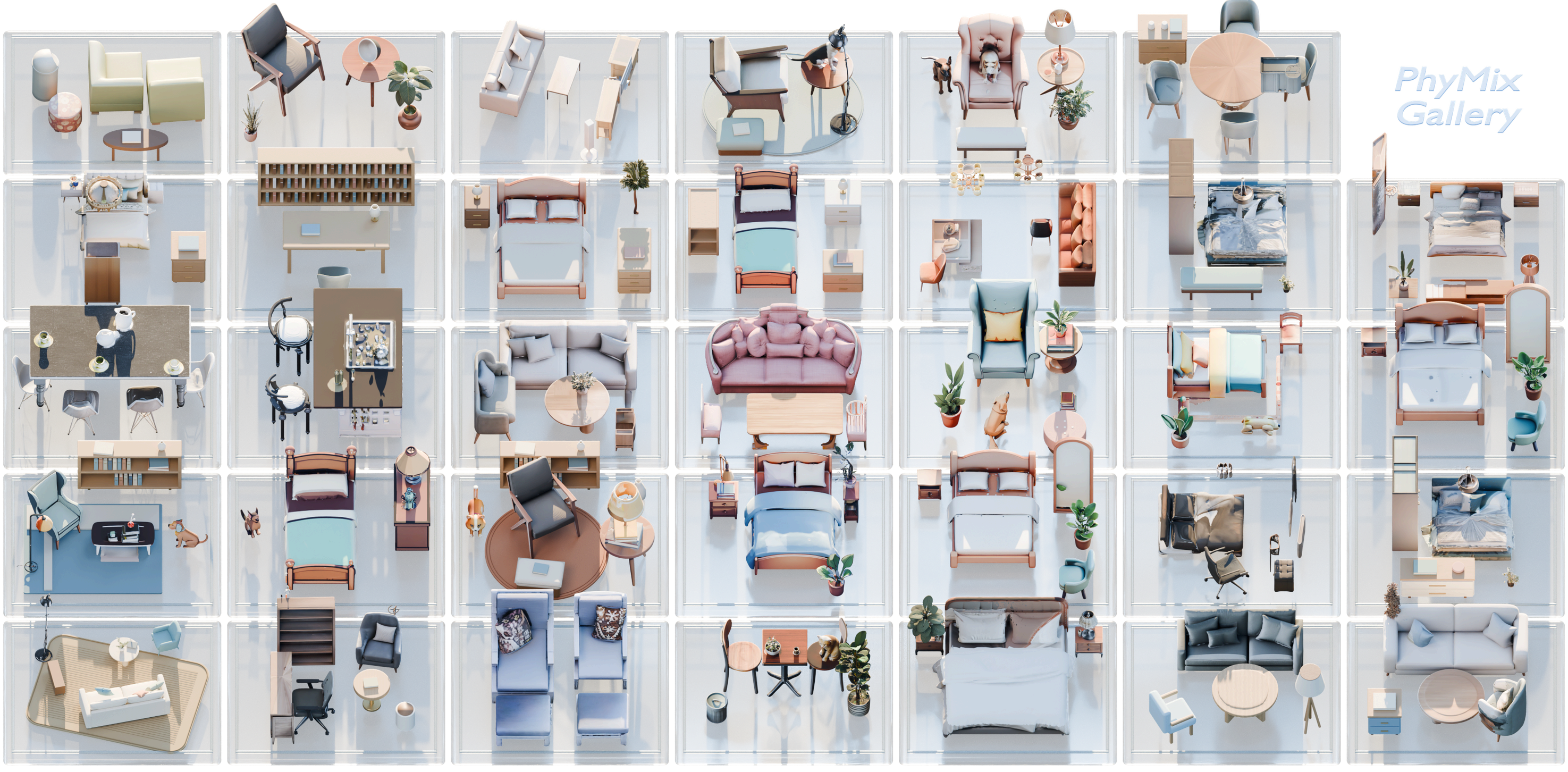}
    \caption{\textbf{PhyMix enables physically consistent 3D indoor scene generation from a single image.}
    Our method produces diverse indoor layouts with reliable object placement, grounding, and spatial organization.
    By integrating physics-aware evaluation and optimization, PhyMix generates scenes that are not only visually faithful but also physically plausible for downstream interaction.}
    \label{fig:intro}
\end{figure*}

\IEEEPARstart{G}{enerating} editable 3D indoor scenes from a single image enables applications in embodied AI, AR/VR design, and large-scale simulation~\cite{long2025survey,wong2025survey}. For applications involving physical interaction, a reconstruction must capture both the observed appearance and a physically plausible arrangement of objects. Interpenetrating furniture, unsupported objects, unstable placements, and obstructed free space can undermine interaction even when the scene looks convincing from the input viewpoint. The central challenge is therefore to recover scenes that respect physical constraints while remaining faithful to the limited visual evidence in a single image.
\begin{table*}[t]
\centering
\setlength{\abovecaptionskip}{2pt}
\setlength{\belowcaptionskip}{3pt}

\caption{Coverage of physical scene properties in existing methods.}
\label{tab:phys-vs-method}

\fontsize{7.2}{7.8}\selectfont
\setlength{\tabcolsep}{2.0pt}
\renewcommand{\arraystretch}{0.82}

\resizebox{0.96\textwidth}{!}{%
\begin{tabular}{l*{9}{c}}
\toprule
& \multicolumn{2}{c}{\textbf{Geometric Priors}}
& \multicolumn{4}{c}{\textbf{Contact}}
& \multicolumn{2}{c}{\textbf{Stability}}
& \textbf{Deployability} \\
\cmidrule(lr){2-3}
\cmidrule(lr){4-7}
\cmidrule(lr){8-9}
\cmidrule(lr){10-10}

\textbf{Method}
& Orientation
& Scale
& Collision
& Grounding
& Support
& Anchoring
& Static
& Dynamic
& Reachability \\
\midrule

CAST~\cite{yao2025cast}
& & & \checkmark & \checkmark & \checkmark & & & & \\

DepR~\cite{zhao2025depr}
& & & & & & & & \checkmark & \\

PhyScene~\cite{yang2024physcene}
& & & \checkmark & \checkmark & & & & & \checkmark \\

RoomCraft~\cite{zhou2025roomcraft}
& \checkmark & & & \checkmark & & & & & \\

HiScene~\cite{dong2025hiscene}
& \checkmark & & & & & & & & \\

Gen3DSR~\cite{ardelean2024gen3dsr}
& & & & & & & & & \\

MIDI~\cite{huang2025midi}
& & & & & & & & & \\

PartCrafter~\cite{lin2025partcrafter}
& & & & & \checkmark & & & & \\

SceneTransporter~\cite{wang2026scenetransporter}
& & & \checkmark & & & & & & \\

SceneGen~\cite{meng2025scenegen}
& & \checkmark & \checkmark & & & & & & \\

CHOrD~\cite{su2025chord}
& \checkmark & & \checkmark & \checkmark & & & & & \\

SAM 3D~\cite{chen2025sam}
& \checkmark & \checkmark & & & & & & & \\

I-Scene~\cite{ling2025scene}
& \checkmark & & & \checkmark & & & & & \\

Scenethesis~\cite{ling2025scenethesis}
& & & \checkmark & \checkmark & & & \checkmark & & \\

SceneSmith~\cite{pfaff2026scenesmith}
& & \checkmark & \checkmark & \checkmark
& \checkmark & & \checkmark & \checkmark & \\

SAGE~\cite{xia2026sage}
& & \checkmark & \checkmark & \checkmark
& \checkmark & & \checkmark & \checkmark & \checkmark \\

PhyScene3D~\cite{chenphyscene3d}
& & & \checkmark & \checkmark
& \checkmark & & \checkmark & & \\

3D-Fixer~\cite{yin20263d}
& & & & & & & & & \\

REST3D~\cite{ma2026rest3d}
& \checkmark & & \checkmark & \checkmark
& \checkmark & \checkmark & \checkmark & \checkmark & \\

PAT3D~\cite{lin2026pat3d}
& & & \checkmark & \checkmark
& \checkmark & & \checkmark & \checkmark & \\

Mira-Scene~\cite{sun2026mira}
& \checkmark & \checkmark & & & & & & & \\

\midrule

\textbf{PhyMix (Ours)}
& \checkmark
& \checkmark
& \checkmark
& \checkmark
& \checkmark
& \checkmark
& \checkmark
& \checkmark
& \checkmark \\

\bottomrule
\end{tabular}%
}

\vspace{1pt}

{\fontsize{6.5}{7}\selectfont
\raggedright
A checkmark indicates an explicit mechanism or diagnostic;
a blank makes no coverage claim.
Coverage does not guarantee physical validity.\par
}

\par\medskip
\begingroup
\normalsize

\centering

\includegraphics[
    width=1\textwidth,
    keepaspectratio
]{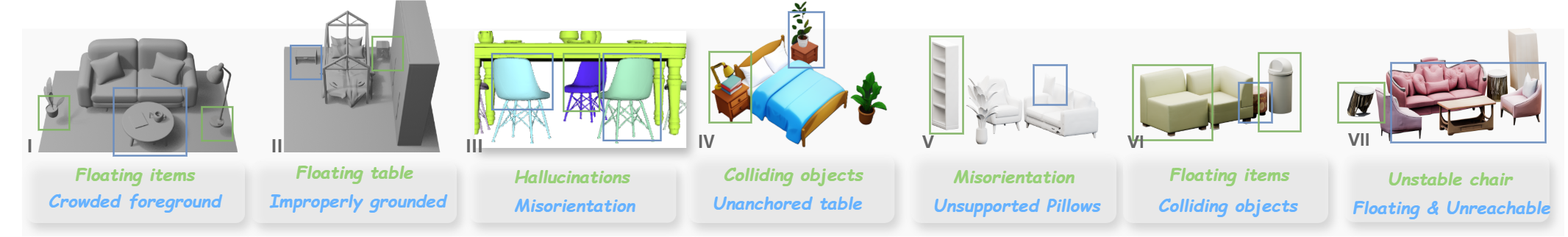}

\captionof{figure}{Representative physical errors in scenes generated by
MIDI (I--II)~\cite{huang2025midi},
PartCrafter (III)~\cite{lin2025partcrafter},
SAM 3D (IV)~\cite{chen2025sam},
PAT3D (V)~\cite{lin2026pat3d},
3D-Fixer (VI)~\cite{yin20263d}, and
REST3D (VII)~\cite{ma2026rest3d}.}
\label{fig:physics-overview}

\endgroup
\end{table*}

Recent progress in single-image scene generation has improved both object geometry and scene composition. Existing approaches include sequential pipelines that reconstruct individual objects and then assemble them~\cite{ardelean2024gen3dsr,han2024reparo,yao2025cast}, joint generators that model object geometry and spatial arrangement together~\cite{huang2025midi,lin2025partcrafter,meng2025scenegen}, and agentic systems that combine asset selection, spatial planning, and iterative tool execution~\cite{ling2025scenethesis,pfaff2026scenesmith,xia2026sage}. Despite these advances, visual agreement with an input image does not ensure physically valid three-dimensional relationships: occlusion can conceal interpenetration, apparent contact need not provide load-bearing support, and a plausible layout may obstruct movement. Some methods incorporate physical checks or optimization, but their coverage varies across constraints and object relationships. This matters because physical constraints are interdependent: resolving a collision may leave an object unsupported or block a passage. These limitations motivate a unified approach that evaluates multiple physical constraints together and uses the resulting feedback to improve both scene generation and subsequent correction.

To this end, we introduce a unified \emph{Physics Evaluator} that assesses nine measurable constraints across four aspects: Geometric Priors, Contact, Stability, and Deployability. It checks these constraints together on each scene and provides two complementary outputs: a scene-level score that summarizes overall physical quality and object-level diagnostics that identify specific violations. Using this evaluator as a benchmark, we find that even state-of-the-art generators produce substantial physical inconsistencies despite their visual plausibility (\cref{tab:benchmark}).

Beyond benchmarking, the evaluator provides feedback on both which scenes are physically better and where an individual scene needs correction. This feedback can support two complementary ways to improve physical consistency. Training can encourage the generator to produce better scenes from the outset, but errors may remain in individual outputs when geometry and support relationships are ambiguous. Repair can address these specific errors, but preserving fidelity to the input image limits how much a generated scene can be changed. A poor initial arrangement may therefore be difficult to fix through local edits alone. These limitations motivate combining training-time alignment with inference-time repair, guided by the same physical criteria.

Building on this idea, we present \textbf{PhyMix}, a framework that combines implicit alignment and explicit refinement (\cref{fig:intro}). For \emph{implicit alignment}, Scene-GRPO samples multiple candidate scenes for each input and uses their scene-level scores to fine-tune the generator toward more physically plausible outputs, with flow-matching losses serving as likelihood proxies. For \emph{explicit refinement}, a verifier-guided repair agent uses object-level diagnostics to propose small edits, such as separating colliding objects, restoring support, correcting orientations, attaching objects to their intended surfaces, and clearing blocked passages. Each proposed edit is tested on a copy of the scene and re-evaluated across the physical constraints to assess both its intended effect and any new violations. The verifier accepts an edit only if it improves physical validity while respecting limits on cumulative changes from the initial reconstruction; object identities remain unchanged. Together, alignment improves the initial scene, and verified repair corrects residual errors.

We evaluate PhyMix on 3D-FRONT~\cite{fu20213d},
InternScenes~\cite{zhong2025internscenes}, and
SAGE-10K~\cite{xia2026sage}. The reported comparisons
show improvements in physical plausibility and reconstruction
quality. On 3D-FRONT, PhyMix with the MIDI backbone reduces
the mean physical violation rate to 1.42\%, compared with
14.74\% for MIDI and 6.80\% for PAT3D, corresponding to
relative reductions of 90.4\% and 79.1\%, respectively.
Component comparisons examine generator alignment and
different repair strategies, while qualitative examples on real-world and stylized images illustrate the framework's
applicability beyond synthetic inputs.

Our contributions are summarized as follows:
\begin{itemize}\setlength{\itemsep}{2pt}\setlength{\parskip}{0pt}\setlength{\parsep}{0pt}
\item A unified Physics Evaluator that organizes nine measurable constraints into four physical aspects and provides scene-level scores for preference learning together with object-level diagnostics for targeted correction.
\item PhyMix, an implicit--explicit framework that couples distribution-level physical alignment with instance-level verified repair, using a shared evaluator to connect learned physical preferences to executable corrections under cumulative content-preservation constraints.
\item Comparative evaluations showing improved physical plausibility and reconstruction quality, together with component analyses and qualitative results on real-world and stylized images.
\end{itemize}
\section{Related Work}
\label{rw}

\subsection{Object-Centric Indoor Scene Generation}
Single-image scene generation combines object geometry recovery with spatial arrangement. Compositional pipelines generate or reconstruct individual objects and align them within a scene~\cite{ardelean2024gen3dsr,han2024reparo,yao2025cast}. Depth-conditioned generation further uses geometric observations to guide object synthesis and layout optimization~\cite{zhao2025depr}, while in-place completion reconstructs objects around spatial anchors extracted from partial scene geometry~\cite{yin20263d}. Joint diffusion models instead couple multiple instances or parts within a shared generation process~\cite{huang2025midi,lin2025partcrafter}. These approaches provide editable scene components and mechanisms for recovering their spatial relationships.

A separate line synthesizes room-scale geometry or extends an input image into navigable environments~\cite{ju2024diffindscene,liang2025wonderland,yu2025wonderworld}. Such representations support visual exploration, whereas object-centric representations expose the individual transforms and relationships needed for targeted physical correction. PhyMix builds on object-centric generators and uses physical feedback to align their output preferences and verify edits to decoded scenes. Its focus is the physical validity of object arrangements while retaining the observed scene content.

\subsection{Physical Evaluation and Optimization}
Physical constraints guide scene generation through collision avoidance, contact reasoning, and simulation. Layout synthesis methods incorporate collision and interactivity guidance~\cite{yang2024physcene}, optimize layouts under floor-plan constraints~\cite{su2025chord}, or introduce physics-guided objectives for compositional generation~\cite{zhou2025layoutdreamer}. For single-image reconstruction, support-tree reasoning and physics-constrained refinement jointly address physical violations and alignment with the input image~\cite{ma2026rest3d}. PAT3D uses hierarchical object relations and a differentiable affine-body simulator to optimize text-conditioned scenes toward stable, non-interpenetrating equilibria~\cite{lin2026pat3d}. Its simulation-in-the-loop optimization differentiates through simulated states to adjust the initial layout. PhyMix instead uses evaluator feedback for preference alignment and candidate verification, without backpropagating through the simulator. These methods establish physical refinement as an important component of scene construction.

Physical feedback has also been incorporated into model alignment. In tabletop generation, PhyScene3D~\cite{chenphyscene3d} combines hierarchical construction with Physics-Aware Denoising Alignment, using SDF-based optimization to obtain physical training signals and refined semantic anchors. PhyMix addresses image-conditioned indoor reconstruction through flow-matching preference alignment and retains a verifier-guided repair stage during inference. Its shared evaluator covers nine constraints and supplies both training preferences and diagnostics for bounded edits, with cumulative constraints on departure from the input reconstruction.

Evaluation frameworks provide a complementary basis for comparing generated scenes. SceneEval~\cite{tam2026sceneeval} assesses explicit requirements and implicit expectations, including support, collision, and navigability. PhyMix builds on this need for interpretable evaluation by defining constraint applicability, aggregation, and object-level diagnostics for an image-conditioned setting. Shared constraint definitions provide failure-rate preferences for alignment and residual-cost feedback for repair.

\subsection{Preference Alignment of Generative Models}
Preference optimization adapts generators using relative output quality. Direct objectives learn from preferred and dispreferred responses~\cite{rafailov2023direct,hong2024orpo}, while Diffusion-DPO extends preference learning to denoising models through a variational likelihood bound~\cite{wallace2024diffusion}. Flow-GRPO~\cite{liu2026flow} and DanceGRPO~\cite{xue2025dancegrpo} instead use stochastic generation trajectories and probability-ratio-based policy updates with group-relative advantages~\cite{shao2024deepseekmath}. Scene-GRPO adopts group-relative physical feedback but uses listwise cross-entropy to match preference targets with a softmax distribution over negative flow-matching errors~\cite{lipman2022flow}. Unlike these policy-gradient methods, it combines generated and perturbed candidates without importance ratios; its scores are finite-candidate likelihood proxies rather than exact scene likelihoods. Verified repair trajectories separately provide imitation and reward-weighted supervision for the action policy, following the principle of learning more strongly from higher-quality actions~\cite{peng2019advantage}.

\subsection{Agentic Scene Construction and Repair}
Agentic systems organize scene construction through planning, asset retrieval or generation, tool execution, and feedback. Image-guided agent pipelines combine scene planning with physics-aware pose optimization~\cite{ling2025scenethesis}; hierarchical agents construct simulator-ready environments with estimated physical properties~\cite{pfaff2026scenesmith}; and visual and physics critics guide iterative object placement and correction~\cite{xia2026sage}. General-purpose agents can also author interactive 3D scenes directly~\cite{openai2026astra}. These systems demonstrate the value of executable actions and feedback, while physical validity remains dependent on the checks applied to the resulting scene.

PhyMix uses agentic actions to repair a fixed image-conditioned reconstruction. Its policy proposes typed, bounded object edits, and its verifier accepts an edit only after checking physical improvement and cumulative content preservation. The distinction lies in coupling this inference-time process with generator alignment under a shared physical evaluator: learned preferences guide scene sampling, and verified actions address residual violations without changing object identities. Accepted trajectories also train the proposal policy, while the verifier remains active during inference.
\section{Are Existing 3D Indoor Scene Generators Physically Reliable?}
\label{sec:reliability}

\begin{table*}[t]
\centering

\caption{Physical consistency and perceptual ratings on 3D-FRONT. Violation rates and severity summaries are percentages (lower is better); MOS uses a 1--7 scale (higher is better). The nine baselines and primary physical rates match \cref{tab:main_physics_results}. Overall Physical Score is $100$ minus the mean of the eight reported violation rates (\cref{eq:benchmark_overall}); severity and MOS are excluded.}
\label{tab:phys_benchmark}

\small
\setlength{\tabcolsep}{3pt}
\renewcommand{\arraystretch}{1.0}

\begin{adjustbox}{max width=\textwidth}
\begin{tabular}{lcccccccccc}
\toprule
Metric & SAM 3D & PartCrafter & DepR & Gen3DSR & REPARO & MIDI & 3D-Fixer & REST3D & PAT3D & \textbf{PhyMix} \\
\midrule

\multicolumn{11}{l}{\emph{Evaluator: Geometric Priors}} \\
Misorientation Rate (\%)
& 8.45 & 5.82 & 11.8 & 14.3 & 14.0 & 7.37 & 9.28 & 4.91 & 6.16 & \textbf{1.72} \\

Scale Violation Rate (\%)
& 7.62 & 4.73 & 9.85 & 12.6 & 11.6 & 6.41 & 7.86 & 5.33 & 6.27 & \textbf{1.82} \\

\midrule

\multicolumn{11}{l}{\emph{Evaluator: Contact}} \\

Collision Rate (\%)
& 8.52 & 3.88 & 9.52 & 16.3 & 13.6 & 7.14 & 10.6 & 8.46 & 2.79 & \textbf{0.56} \\

Collision Severity (\%)
& 3.9 & 3.16 & 4.37 & 7.76 & 5.74 & 2.72 & 4.9 & 3.9 & 0.4 & \textbf{0.05} \\

Floating Rate (\%)*
& 35.6 & 28.4 & 37.1 & 45.4 & 42.5 & 33.2 & 36.8 & 12.7 & 5.44 & \textbf{0.97} \\

Floating Gap Severity (\%)*
& 30.7 & 23.28 & 32.73 & 36.47 & 36.60 & 29.90 & 31.7 & 10.9 & 1.3 & \textbf{1.10} \\

Unanchored Rate (\%)
& 14.8 & 9.24 & 19.5 & 28.2 & 25.3 & 12.8 & 16.4 & 7.08 & 9.38 & \textbf{1.80} \\

\midrule

\multicolumn{11}{l}{\emph{Evaluator: Stability}} \\

Static Instability Rate (\%)
& 26.5 & 15.8 & 34.4 & 37.8 & 37.0 & 22.6 & 27.9 & 9.64 & 6.73 & \textbf{1.27} \\

Dynamic Instability Rate (\%)
& 28.2 & 18.9 & 37.0 & 41.3 & 39.7 & 24.6 & 30.5 & 11.2 & 10.21 & \textbf{2.20} \\

\midrule

\multicolumn{11}{l}{\emph{Evaluator: Deployability}} \\

Unreachable Rate (\%)
& 4.90 & 2.94 & 6.51 & 8.67 & 7.89 & 3.78 & 5.72 & 4.48 & 7.45 & \textbf{1.01} \\

\midrule

\textbf{Overall Physical Score}
& 83.2 & 88.8 & 79.3 & 74.4 & 76.1 & 85.3 & 81.9 & 92.0 & 93.2 & \textbf{98.6} \\

\midrule

\multicolumn{11}{l}{\emph{User Study $\dagger$}} \\

Contact MOS
& 4.0 & 4.6 & 3.8 & 3.1 & 3.4 & 4.2 & 4.0 & 5.4 & 6.8 & \textbf{6.9} \\

Stability MOS
& 3.8 & 4.8 & 3.5 & 2.9 & 3.2 & 4.1 & 3.8 & 5.5 & 6.7 & \textbf{6.8} \\

Geometric Prior MOS
& 4.5 & 5.1 & 4.2 & 3.6 & 3.9 & 4.7 & 4.5 & 5.7 & 6.8 & \textbf{6.9} \\

\textbf{Overall MOS}
& 4.1 & 4.8 & 3.8 & 3.2 & 3.5 & 4.3 & 4.1 & 5.5 & 6.8 & \textbf{6.87} \\

\midrule

\multicolumn{11}{l}{\emph{Pairwise Preference (win rate, \%)}} \\

PhyMix vs.\ PartCrafter
& \multicolumn{10}{c}{92 \,(N{=}200)} \\

PhyMix vs.\ MIDI
& \multicolumn{10}{c}{96 \,(N{=}200)} \\

PhyMix vs.\ REST3D
& \multicolumn{10}{c}{97 \,(N{=}200)} \\

PhyMix vs.\ PAT3D
& \multicolumn{10}{c}{97 \,(N{=}200)} \\

\bottomrule
\end{tabular}
\end{adjustbox}

\smallskip

\small
*\emph{``Floating'' aggregates grounding and support violations.}
$\dagger$ \emph{Higher MOS indicates better perceptual quality. Preference values represent the scene-macro win rate, where ties are counted as 0.5.}
\label{tab:benchmark}
\end{table*}

\subsection{Representative Baselines}
\label{sec:baselines}
To assess whether existing methods generate physically plausible scenes, we evaluate nine representative methods: SAM 3D~\cite{chen2025sam}, PartCrafter~\cite{lin2025partcrafter}, DepR~\cite{zhao2025depr}, Gen3DSR~\cite{ardelean2024gen3dsr}, REPARO~\cite{han2024reparo}, MIDI~\cite{huang2025midi}, 3D-Fixer~\cite{yin20263d}, REST3D~\cite{ma2026rest3d}, and PAT3D~\cite{lin2026pat3d}. All methods are adapted to a common single-image and instance-mask input setting (\suppsec{app:baselines}).

\Cref{tab:phys-vs-method} summarizes the physical properties explicitly addressed by existing methods, while \cref{tab:phys_benchmark} reports the measured physical violation rates of the nine baselines on 3D-FRONT, alongside additional severity and perceptual measurements where available. PhyMix results are included for comparison.

\subsection{Unified Physics Evaluator}
\label{sec:evaluator}
To assess these methods under consistent physical criteria, we introduce a unified \emph{Physics Evaluator} that checks both individual objects and their relationships within a scene. It combines geometric checks with physical simulation to identify violations that may not be apparent in rendered images, such as interpenetration, missing support, and unstable placements (\cref{fig:physics-overview}). The evaluator covers nine measurable constraints across four aspects---Geometric Priors, Contact, Stability, and Deployability---and provides a scene-level score for comparison across methods, together with object-level diagnostics for targeted repair. We describe these four aspects below.

\noindent\textbf{Geometric Priors.}
Orientation and scale measure whether objects conform to category-specific upright directions and plausible dimensions. These category priors complement geometric and mechanical checks: an unusual size is a prior violation, while physical stability is assessed separately.

\noindent\textbf{Contact.}
Collision tests detect interpenetration between objects and with the room.
Grounding checks contact with the floor, support checks contact with an assigned
supporting object, and anchoring checks explicitly required wall or ceiling
attachments. Each object is checked by only one of grounding and support,
so the benchmark tables report the two together as a single Floating column.
Both tests fail an object that floats above or sinks into its support.
A freestanding object is not required to touch a wall unless an attachment
relation is specified.

\noindent\textbf{Stability.}
Static stability tests whether the projected center of mass lies within the support-contact hull under a uniform-density rigid-body assumption, including the combined load of supported objects. Mounted objects are evaluated through their attachment constraints. Contact and stability describe different properties: touching a surface does not ensure equilibrium. Dynamic stability tests pose changes and residual motion during short rigid-body rollouts in MuJoCo, using compound collision shapes obtained by convex decomposition of object meshes. These measurements distinguish making contact from remaining in a stable arrangement under gravity and controlled perturbations.

\noindent\textbf{Deployability.}
Reachability measures whether fixed start--goal pairs remain connected through free space after accounting for the agent's dimensions. This captures obstruction of movement that object-level contact and stability checks alone do not measure.

Each constraint is evaluated only on the objects or relations to which it applies, using the same applicability rules across methods. We report the fraction of eligible objects, relations, or start--goal pairs that violate each constraint, together with measures of violation severity. The evaluator also identifies the affected objects and relations to guide targeted repair. Detailed definitions and thresholds are given in \suppsec{sec:subevaluator}. \Cref{sec:grpo} uses the scene-level scores to align the generator, and \cref{sec:repair} uses the object-level diagnostics to guide repair.
\subsection{Empirical Benchmarking of Physical Consistency}
\label{sec:benchmarking}
\Cref{tab:phys_benchmark} evaluates nine representative methods on 3D-FRONT and reveals persistent physical violations in their generated scenes. For example, PartCrafter exhibits floating and dynamic instability rates of 28.4\% and 18.9\%, respectively. Methods that explicitly incorporate physical constraints reduce several types of violations, but errors remain in contact, support, stability, and free-space accessibility. These results highlight the need for further improvements in physical consistency across different generation strategies.

The user study complements these measurements with human judgments of geometric plausibility, contact, and perceived stability. Thirty participants rate rendered scenes on a 1--7 scale and compare paired outputs with randomized presentation (\suppsec{US}). PhyMix achieves an Overall MOS of 6.87, compared with 3.2--6.8 for the nine baselines, and is preferred over PartCrafter, MIDI, REST3D, and PAT3D in 92\%, 96\%, 97\%, and 97\% of paired comparisons, respectively (200 comparisons per baseline). These results provide evidence beyond the evaluator scores that the generated scenes also appear more plausible to human observers.

The remaining violations motivate using the evaluator not only to compare methods, but also to improve their outputs. Its scene-level scores can guide the generator toward more physically plausible scenes, while its object-level diagnostics can direct corrections to specific errors. The following section introduces PhyMix, which combines training-time alignment with verified inference-time repair using these two forms of feedback.
\FloatBarrier

\section{PhyMix}
\label{sec:method}

\begin{figure*}[t]
\centering
\includegraphics[width=0.95\textwidth]{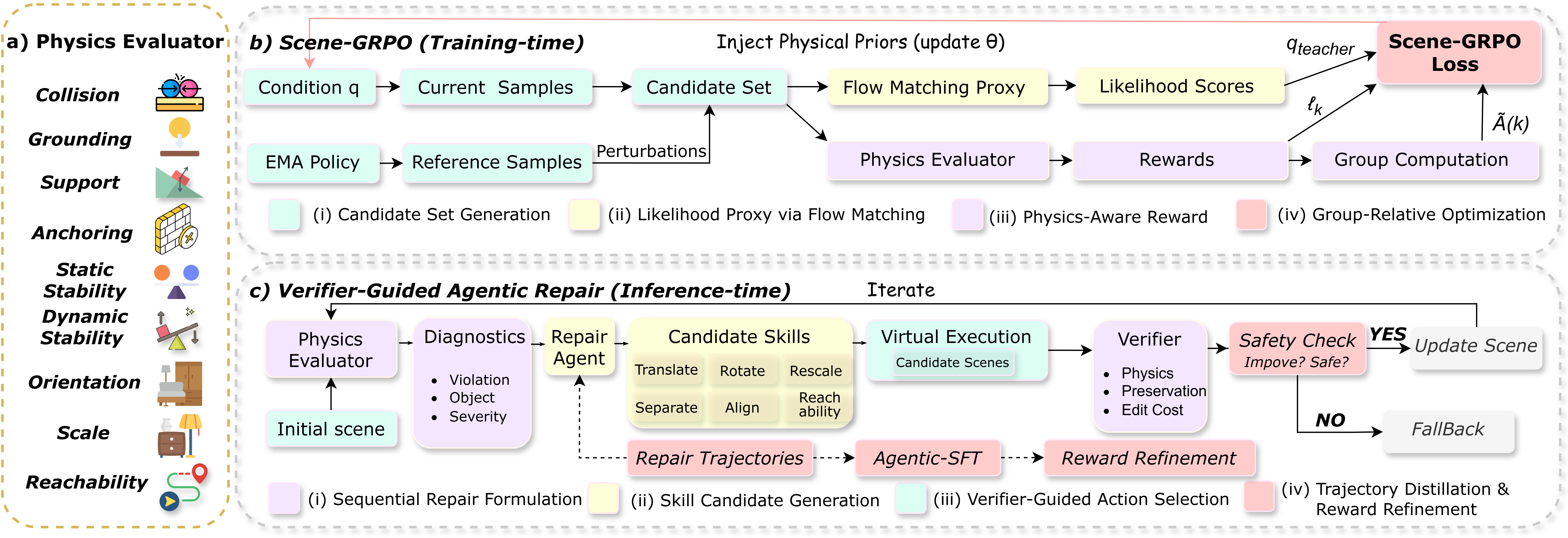}
\caption{\textbf{Overview of PhyMix.} (a) The Physics Evaluator measures geometric priors, contact, stability, and deployability. (b) Scene-GRPO aligns the generator through candidate generation, flow-matching likelihood scoring, physical preferences, and group-relative optimization. (c) Verifier-guided agentic repair corrects residual violations through diagnosis, skill proposals, testing edits on scene copies, and verified updates. Committed edits and candidate verification outcomes supervise the local repair policy.}
\label{fig:overview}
\end{figure*}

Given an RGB image and instance masks $X=(I,\mathcal M)$, a pretrained object-level generator $\pi_\theta(S\mid X)$, such as MIDI~\cite{huang2025midi} or PartCrafter~\cite{lin2025partcrafter}, produces a scene
\begin{equation}
 S=\{(z_i,\psi_i,c_i)\}_{i=1}^N,\qquad \psi_i=(t_i,q_i,s_i),
 \label{eq:scene_representation}
\end{equation}
where $z_i$ is the geometry of object $i$, $c_i$ its category, and $\psi_i$ its pose, consisting of a translation $t_i\in\R^3$, a unit quaternion $q_i$, and a scale $s_i\in\R_+^3$. The camera, the object identities, and the support relation of each object are estimated once from the input (\suppsec{app:eval_scope}).

PhyMix improves this generator in two complementary ways (\cref{fig:overview}):
\begin{equation}
 S_0\sim\pi_{\hat\theta}(\cdot\mid X),\qquad
 S_{\rm out}=\mathcal A_{\rm repair}(S_0;X,\mathcal V).
 \label{eq:overall_obj_v2}
\end{equation}
During training, Scene-GRPO (\cref{sec:grpo}) uses the scene-level scores of the Physics Evaluator $\mathcal V$ to align the generator, so that the aligned parameters $\hat\theta$ produce more physically plausible initial scenes $S_0$. At inference, verifier-guided repair (\cref{sec:repair}) uses the evaluator's object-level diagnostics to correct the violations that remain in $S_0$.

\subsection{Scene-GRPO: Training-Time Implicit Alignment}
\label{sec:grpo}
Scene-GRPO compares a group of candidate scenes for each input (\cref{fig:overview}(b)). The Physics Evaluator determines which candidates are better, and training makes the generator assign higher scores to them. Because rewards are compared only within a group, no value network is needed.

\medskip
\noindent\textbf{Candidate group.}
For each input $X$, we form $K$ candidates $\{S^{(k)}\}_{k=1}^K$. Half are sampled from the current generator. The other half are sampled from an exponential moving average (EMA) of the generator and then slightly perturbed in object position, orientation, and scale, which widens the range of physical quality within the group. Each candidate has a reference scene $R^{(k)}$ for measuring layout change: the unperturbed sample for a perturbed candidate, and an EMA sample with a fixed seed for a current-model candidate.

\medskip
\noindent\textbf{Model scores.}
A flow-matching generator~\cite{lipman2022flow} has no tractable likelihood, so we score each candidate by its flow-matching loss
\begin{equation}
 \ell_\theta^{(k)}=\mathbb E_{t,\epsilon}
 [\mathcal L_{\rm FM}(\theta;S^{(k)},X,t,\epsilon)],
 \label{eq:fm_candidate}
\end{equation}
where the noise level $t$ and the noise $\epsilon$ are shared by all candidates in a group (\suppsec{app:ref_lik}). A low loss indicates that the candidate is close to what the generator tends to produce. Normalizing over the group gives the model scores
\begin{equation}
 p_\theta^{(k)}=
 \frac{\exp(-\ell_\theta^{(k)}/\tau_p)}
 {\sum_m\exp(-\ell_\theta^{(m)}/\tau_p)},
 \label{eq:model_distribution}
\end{equation}
which rank candidates within the group rather than define exact scene likelihoods. The EMA model computes the same scores, giving a reference distribution $p_{\rm ref}$.

\medskip
\noindent\textbf{Physical preferences.}
\label{sec:physics_feedback}
The Physics Evaluator measures the violation rate of each applicable constraint and averages these rates into $P(S)\in[0,1]$ (\suppsec{app:feedback_aggregation}). The corresponding physical score is
\begin{equation}
 Q_9(S)=100\bigl(1-P(S)\bigr).
 \label{eq:phys_score}
\end{equation}
To prevent the generator from avoiding violations by drifting away from the input layout, the reward also penalizes layout change:
\begin{equation}
 r_k=-P(S^{(k)})
 -\lambda_{\rm align}D_{\rm lay}(S^{(k)},R^{(k)}),
 \label{eq:reward}
\end{equation}
where $D_{\rm lay}$ measures the change in object position, orientation, and scale (\suppsec{app:repair_utility}). The rewards are normalized within the group and clipped,
\begin{equation}
 A_k=\frac{r_k-\bar r}{\sigma_r+\varepsilon},\qquad
 \bar A_k=\clip(A_k,-a,a),
\end{equation}
where $\bar r$ and $\sigma_r$ are the group mean and standard deviation, and then converted into a target distribution
\begin{equation}
 q_k=
 \frac{\exp(\bar A_k/\tau_r)}
 {\sum_m\exp(\bar A_m/\tau_r)},
 \label{eq:physics_distribution}
\end{equation}
in which better candidates receive larger weights.

\medskip
\noindent\textbf{Optimization.}
The generator is trained to move its model scores toward the target:
\begin{equation}
\begin{split}
 \mathcal L_{\rm GRPO}={}&
 -\sum_k q_k\log p_\theta^{(k)}
 +\beta_u\operatorname{KL}(p_{\rm ref}\|p_\theta)\\
 &+\lambda_{\rm FM}\mathcal L_{\rm anchor}.
\end{split}
\label{eq:grpo}
\end{equation}
The first term aligns the model scores with the physical preferences. The KL term, whose weight $\beta_u$ decays during training, keeps the scores close to those of the EMA model for stable updates. The anchor term $\mathcal L_{\rm anchor}$ is the original flow-matching loss on training scenes and preserves reconstruction quality. The targets $q$ and $p_{\rm ref}$ are fixed within each update, and groups with near-identical rewards use only the anchor term. We update the generator with LoRA~\cite{hu2022lora} while freezing its geometry encoder and decoder (\suppsec{app:impl}).

\subsection{Verifier-Guided Agentic Repair}
\label{sec:repair}
Individual scenes from the aligned generator can still contain errors, such as a lamp floating above a table. Repair corrects them with an agent that edits the scene step by step (\cref{fig:overview}(c)): a policy proposes edits, and a verifier commits an edit only if it improves physical consistency while keeping the scene close to the input. We first describe one repair step and then the training of the policy.

\medskip
\noindent\textbf{Observation and actions.}
At step $\ell$, the evaluator returns object-level diagnostics $\mathcal D_\ell=\mathcal V(S_\ell)$ for the current scene: which constraints fail, on which objects or relations, and by how much. The observation $o_\ell$ combines these diagnostics with the input image and masks, a rendering of $S_\ell$, the object transforms, the previous edits, and the remaining budget. The policy acts through seven predefined skills, each targeting one type of error: translate, rotate, rescale, separate collision, snap to support, align to wall, and reachability adjustment (\suppsec{app:skills}). Each step tests up to $M$ edits,
\begin{equation}
 \mathcal A_\ell=\{a^{(m)}_\ell\}_{m=1}^{M},\qquad a=(\kappa,i,j,\vartheta),
 \label{eq:repair_action}
\end{equation}
where $\kappa$ is the skill, $i$ the target object, $j$ an optional related object or surface, and $\vartheta$ the bounded edit parameters. The policy can also output STOP.

\medskip
\noindent\textbf{Verification.}
Each edit is executed on a copy of the scene, $\tilde S^{(m)}=u_{a^{(m)}_\ell}(S_\ell)$, and all constraints are re-evaluated, since fixing one error may create another. The verifier scores each copy by
\begin{equation}
 U(S)=-\mathcal C(S)-\lambda_{\rm pres}D(S,S_0;X),
 \label{eq:repair_utility}
\end{equation}
where $\mathcal C$ is a graded violation cost that decreases as errors shrink, even before a test passes (\suppsec{app:repair_cost}), and $D$ measures the layout and mask deviation from the initial scene $S_0$. The net gain of an edit also subtracts its size $c(a;N)$:
\begin{equation}
 \Delta J_m=U(\tilde S^{(m)})-U(S_\ell)
 -\lambda_{\rm cost}\,c(a^{(m)}_\ell;N).
 \label{eq:repair_gain}
\end{equation}
An edit passes verification if it reduces $\mathcal C$, respects the editing and cumulative preservation limits, and creates or worsens no severe violation. Among the passing edits $\mathcal F_\ell$, the verifier selects
\begin{equation}
 m^\star=\arg\max_{m\in\mathcal F_\ell}\Delta J_m
 \label{eq:repair_select}
\end{equation}
and commits it if $\Delta J_{m^\star}>\delta$. If no edit is committed, a rule-based fallback is tried once. Repair stops when all constraints pass, when no edit can be committed, or when the budget is exhausted (\suppsec{app:repair_utility}). Because every scene is compared with $S_0$ and only verified edits are committed, repair stays close to the input and a poor proposal is simply rejected.

\medskip
\noindent\textbf{Trajectories and local policy.}
We collect repair trajectories on training rooms with GPT-5.6-Sol as the policy, which writes each edit as a JSON action. At every step, a trajectory records the observation, the proposed edits with their verification results, and the committed edit, or STOP at the end. Because querying GPT-5.6-Sol at every step is slow, we distill these trajectories into a local policy based on Qwen3.5-9B. Instead of generating text, the local policy predicts each field of an edit with a structured output head: classification heads for the skill, the objects, and STOP, and a regression head for the parameters. An additional head outputs an acceptance score $s_\psi(o,a)$, which predicts whether an edit will pass verification. At inference, the policy proposes one edit per step, and rule-based proposals fill the remaining slots (\suppsec{app:impl}).

\medskip
\noindent\textbf{Supervised fine-tuning.}
Since each decision depends on the scene produced by the previous one, the policy is trained one step at a time: each example pairs an observation with the committed edit, or with STOP at the final step. The loss is
\begin{equation}
 \mathcal L_{\rm SFT}=\mathcal L_{\rm skill}+\mathcal L_{\rm target}
 +\mathcal L_{\rm param}+\mathcal L_{\rm stop}+\mathcal L_{\rm accept},
 \label{eq:repair_sft}
\end{equation}
where $\mathcal L_{\rm skill}$, $\mathcal L_{\rm target}$, and $\mathcal L_{\rm stop}$ are classification losses; $\mathcal L_{\rm param}$ is a fixed-variance Gaussian negative log-likelihood, i.e., a scaled squared error, applied only to the fields used by the chosen skill; and $\mathcal L_{\rm accept}$ is a binary loss on all proposed edits, labeled by whether they passed verification. This stage gives the \emph{Agentic-SFT} policy.

\medskip
\noindent\textbf{Reward refinement.}
SFT weights all committed edits equally and does not compare edits proposed for the same scene. Reward refinement adds two terms that exploit the verifier's gains. A ranking loss prefers an edit $a^+$ that passes verification over a rejected edit $a^-$, or over a passing edit whose net gain is lower by at least $\delta$:
\begin{equation}
 \mathcal L_{\rm rank}=-\log\sigma\bigl(s_\psi(o,a^+)-s_\psi(o,a^-)\bigr),
 \label{eq:repair_rank}
\end{equation}
where $\sigma$ is the sigmoid function. Following advantage-weighted regression~\cite{peng2019advantage}, a weighted imitation loss emphasizes committed edits that remove more error:
\begin{equation}
 \mathcal L_{\rm rw}=-\mathbb E_{\rm committed}\bigl[w(g)\log\rho_\psi(a\mid o)\bigr],
 \label{eq:repair_rw}
\end{equation}
where $\rho_\psi(a\mid o)$ is the policy's likelihood of the committed edit, $g=\mathcal C(S_\ell)-\mathcal C(S_{\ell+1})$ is its cost reduction, and $w(g)$ increases with $g$ up to a cap (\suppsec{app:impl}). The final objective is
\begin{equation}
 \mathcal L_{\rm repair}=\mathcal L_{\rm SFT}+\mu\mathcal L_{\rm rank}+\nu\mathcal L_{\rm rw},
 \label{eq:repair_total}
\end{equation}
which gives the \emph{Agentic-SFT+Reward} policy used at inference. Every edit it proposes is still checked by the verifier.
\section{Experiments}
\subsection{Experimental Settings}

\noindent\textbf{Datasets and comparisons.}
We train and evaluate PhyMix on 3D-FRONT~\cite{fu20213d},
using the MIDI preprocessing pipeline~\cite{huang2025midi}
with 12K training scenes and 4.8K test scenes.
We also test the trained model on
InternScenes~\cite{zhong2025internscenes} and
SAGE-10K~\cite{xia2026sage} without further fine-tuning
to assess generalization across datasets.

Our main experiments use MIDI as the scene generator.
We additionally apply PhyMix to PartCrafter to test its
compatibility with another backbone. PAT3D~\cite{lin2026pat3d} is adapted to image input by skipping its text-to-image stage (\suppsec{app:baselines}), and we apply
only the repair stage and leave the generator unchanged. 
We compare against the nine methods introduced in
\cref{sec:baselines}, using the same RGB images and instance
masks as inputs (\suppsec{app:baselines}).
We report three versions of PhyMix:
\emph{GRPO only}, which uses generator alignment without repair;
\emph{GRPO + SFT}, which adds an SFT-trained repair policy;
and \emph{GRPO + SFT + Reward}, the full framework with
a reward-refined repair policy.

\medskip
\noindent\textbf{Implementation.}
We use GPT-5.6-Sol to propose edits when collecting repair
trajectories, and train Qwen3.5-9B on these trajectories
as the local repair policy.
Generator alignment uses the same LoRA settings for
MIDI and PartCrafter.
At inference, we limit the number of proposed edits,
accepted edits, and evaluator calls per scene.
Training settings and repair budgets are provided in
\suppsec{app:impl} and \suppsec{app:repair_utility}.

\medskip
\noindent\textbf{Evaluation metrics.}
We report physical violation rates as percentages;
lower values indicate fewer errors.
For geometric reconstruction, we report scene-level Chamfer distance (CD-S) and F-score (F-Score-S),
as well as bounding-box IoU.

The benchmark tables contain eight physical violation columns.
Grounding and support are combined into a single
\emph{Floating} column; the other seven constraints are
reported separately.
We summarize these results with an overall score:
\begin{equation}
 \mathrm{Overall}_{\rm rep}
 =100-\frac{1}{8}\sum_{j=1}^{8}e_j,
 \label{eq:benchmark_overall}
\end{equation}
where $e_j$ is the percentage violation rate in column $j$.
A higher overall score indicates better physical consistency.
This table score gives equal weight to the eight reported
columns. It differs from the training score $Q_9$, which
treats grounding and support separately and averages over
the applicable constraints within each scene.
We report overall scores to three significant digits;
dashes indicate measurements that are not reported.

\subsection{Comparison with Existing Methods}
\label{sec:main_results}
\begin{table*}[!t]
\centering
\caption{Physical-consistency comparison on three datasets. All baselines use adapted RGB-image and instance-mask inputs. PhyMix is trained on 3D-FRONT without dataset-specific fine-tuning on the other two datasets. Violation rates are percentages; Overall rep is defined in \cref{eq:benchmark_overall}.}
\label{tab:main_physics_results}
\footnotesize
\setlength{\tabcolsep}{2.8pt}
\begin{adjustbox}{max width=\linewidth}
\begin{tabular}{llccccccccc}
\toprule
Dataset & Method & Collision$\downarrow$ & Floating$\downarrow$ & Unanchored$\downarrow$ & Static Inst.$\downarrow$ & Dynamic Inst.$\downarrow$ & Misori.$\downarrow$ & Scale Viol.$\downarrow$ & Unreach.$\downarrow$ & $\mathrm{Overall}_{\rm rep}\uparrow$ \\
\midrule

\multirow{12}{*}{3D-FRONT}
& SAM 3D~\cite{chen2025sam} & 8.52 & 35.6 & 14.8 & 26.5 & 28.2 & 8.45 & 7.62 & 4.90 & 83.2 \\
& PartCrafter~\cite{lin2025partcrafter} & 3.88 & 28.4 & 9.24 & 15.8 & 18.9 & 5.82 & 4.73 & 2.94 & 88.8 \\
& DepR~\cite{zhao2025depr} & 9.52 & 37.1 & 19.5 & 34.4 & 37.0 & 11.8 & 9.85 & 6.51 & 79.3 \\
& Gen3DSR~\cite{ardelean2024gen3dsr} & 16.3 & 45.4 & 28.2 & 37.8 & 41.3 & 14.3 & 12.6 & 8.67 & 74.4 \\
& REPARO~\cite{han2024reparo} & 13.6 & 42.5 & 25.3 & 37.0 & 39.7 & 14.0 & 11.6 & 7.89 & 76.1 \\
& MIDI~\cite{huang2025midi} & 7.14 & 33.2 & 12.8 & 22.6 & 24.6 & 7.37 & 6.41 & 3.78 & 85.3 \\
& 3D-Fixer~\cite{yin20263d} & 10.6 & 36.8 & 16.4 & 27.9 & 30.5 & 9.28 & 7.86 & 5.72 & 81.9 \\
& REST3D~\cite{ma2026rest3d} & 8.46 & 12.7 & 7.08 & 9.64 & 11.2 & 4.91 & 5.33 & 4.48 & 92.0 \\
& PAT3D~\cite{lin2026pat3d} & 2.79 & 5.44 & 9.38 & 6.73 & 10.2 & 6.16 & 6.27 & 7.45 & 93.2 \\
\cmidrule(lr){2-11}
& GRPO only & 2.32 & 6.73 & 3.85 & 3.33 & 4.15 & 2.18 & 1.95 & 1.42 & 96.8 \\
& GRPO + SFT & 0.840 & 1.68 & 2.35 & 1.85 & 2.95 & 1.86 & 1.92 & 1.18 & 98.2 \\
& GRPO + SFT + Reward
& \textbf{0.560}
& \textbf{0.970}
& \textbf{1.80}
& \textbf{1.27}
& \textbf{2.20}
& \textbf{1.72}
& \textbf{1.82}
& \textbf{1.01}
& \textbf{98.6} \\
\midrule

\multirow{12}{*}{InternScenes}
& SAM 3D~\cite{chen2025sam} & 9.24 & 41.2 & 16.5 & 28.3 & 29.8 & 9.35 & 8.42 & 5.65 & 81.4 \\
& PartCrafter~\cite{lin2025partcrafter} & 5.10 & 34.6 & 11.8 & 19.4 & 22.1 & 7.10 & 6.35 & 3.62 & 86.2 \\
& DepR~\cite{zhao2025depr} & 10.4 & 39.8 & 21.2 & 35.6 & 38.4 & 12.6 & 10.5 & 6.98 & 78.1 \\
& Gen3DSR~\cite{ardelean2024gen3dsr} & 17.8 & 48.6 & 30.4 & 39.5 & 43.0 & 15.1 & 13.4 & 9.20 & 72.9 \\
& REPARO~\cite{han2024reparo} & 14.9 & 45.3 & 27.1 & 38.2 & 41.5 & 14.7 & 12.5 & 8.44 & 74.7 \\
& MIDI~\cite{huang2025midi} & 7.85 & 38.5 & 14.2 & 24.1 & 26.2 & 8.12 & 7.15 & 4.35 & 83.7 \\
& 3D-Fixer~\cite{yin20263d} & 12.3 & 44.2 & 19.1 & 31.8 & 34.6 & 10.9 & 9.35 & 7.02 & 78.8 \\
& REST3D~\cite{ma2026rest3d} & 10.4 & 19.3 & 9.72 & 14.2 & 16.1 & 6.28 & 6.74 & 6.18 & 88.9 \\
& PAT3D~\cite{lin2026pat3d} & 6.94 & 13.2 & 11.3 & 9.87 & 13.0 & 10.2 & 10.6 & 8.22 & 89.6 \\
\cmidrule(lr){2-11}
& GRPO only & 3.10 & 8.45 & 5.20 & 4.42 & 5.75 & 4.12 & 4.05 & 2.83 & 95.3 \\
& GRPO + SFT & 2.45 & 5.10 & 4.56 & 3.78 & 5.08 & 3.82 & 3.78 & 2.62 & 96.1 \\
& GRPO + SFT + Reward
& \textbf{2.15}
& \textbf{3.82}
& \textbf{4.15}
& \textbf{3.24}
& \textbf{4.56}
& \textbf{3.55}
& \textbf{3.62}
& \textbf{2.45}
& \textbf{96.6} \\
\midrule

\multirow{12}{*}{SAGE-10K}
& SAM 3D~\cite{chen2025sam} & 9.76 & 43.8 & 17.2 & 30.1 & 31.7 & 9.88 & 8.95 & 6.10 & 80.3 \\
& PartCrafter~\cite{lin2025partcrafter} & 5.65 & 36.9 & 12.7 & 21.0 & 23.4 & 7.68 & 6.90 & 4.10 & 85.2 \\
& DepR~\cite{zhao2025depr} & 10.8 & 42.5 & 22.0 & 36.7 & 39.3 & 13.1 & 11.2 & 7.35 & 77.1 \\
& Gen3DSR~\cite{ardelean2024gen3dsr} & 18.5 & 50.2 & 31.5 & 40.8 & 44.1 & 15.8 & 14.1 & 9.65 & 71.9 \\
& REPARO~\cite{han2024reparo} & 15.8 & 47.1 & 28.4 & 39.6 & 42.7 & 15.2 & 13.0 & 8.92 & 73.7 \\
& MIDI~\cite{huang2025midi} & 8.32 & 40.5 & 15.6 & 25.8 & 27.6 & 8.76 & 7.82 & 4.94 & 82.6 \\
& 3D-Fixer~\cite{yin20263d} & 13.2 & 47.2 & 20.6 & 34.1 & 36.8 & 11.8 & 10.2 & 7.72 & 77.3 \\
& REST3D~\cite{ma2026rest3d} & 11.3 & 22.8 & 10.8 & 15.8 & 17.8 & 6.95 & 7.42 & 6.92 & 87.5 \\
& PAT3D~\cite{lin2026pat3d} & 7.56 & 15.3 & 12.3 & 11.2 & 14.3 & 11.0 & 11.5 & 11.5 & 88.2 \\
\cmidrule(lr){2-11}
& GRPO only & 3.80 & 9.60 & 5.88 & 5.02 & 6.36 & 4.78 & 4.65 & 3.25 & 94.6 \\
& GRPO + SFT & 3.05 & 6.20 & 5.22 & 4.46 & 5.82 & 4.42 & 4.48 & 3.02 & 95.4 \\
& GRPO + SFT + Reward
& \textbf{2.68}
& \textbf{4.96}
& \textbf{4.78}
& \textbf{4.10}
& \textbf{5.35}
& \textbf{4.18}
& \textbf{4.30}
& \textbf{2.86}
& \textbf{95.8} \\
\bottomrule
\end{tabular}
\end{adjustbox}
\end{table*}

\noindent\textbf{Physical consistency.}
\Cref{tab:main_physics_results} shows that PhyMix achieves
the highest overall physical score on all three datasets.
On 3D-FRONT, it reaches 98.6, compared with 85.3 for MIDI
and 93.2 for the strongest external baseline, PAT3D.
It also leads on InternScenes and SAGE-10K without additional
fine-tuning, supporting generalization across datasets.
Alignment alone raises MIDI's score to 96.8, while repair
further reduces the remaining collisions, floating objects,
and instability.

\begin{table}[hbt]
\centering
\caption{Geometry reconstruction on 3D-FRONT under our unified evaluation protocol. We report scene-level Chamfer distance (CD-S), F-score (F-Score-S), and oriented bounding-box IoU (IoU-B).}
\label{tab:quality_results}
\footnotesize
\setlength{\tabcolsep}{4pt}
\begin{adjustbox}{max width=\linewidth}
\begin{tabular}{lccc}
\toprule
Method
& CD-S$\downarrow$
& F-Score-S$\uparrow$
& IoU-B$\uparrow$ \\
\midrule
SAM 3D~\cite{chen2025sam}
& 0.052 & 61.2 & 0.63 \\
PartCrafter~\cite{lin2025partcrafter}
& 0.098 & 55.9 & -- \\
DepR~\cite{zhao2025depr}
& 0.104 & 52.0 & 0.57 \\
Gen3DSR~\cite{ardelean2024gen3dsr}
& 0.103 & 41.7 & 0.41 \\
REPARO~\cite{han2024reparo}
& 0.109 & 43.5 & 0.40 \\
MIDI~\cite{huang2025midi}
& 0.072 & 57.2 & 0.60 \\
3D-Fixer~\cite{yin20263d}
& 0.096 & 53.6 & 0.54 \\
REST3D~\cite{ma2026rest3d}
& 0.051 & 67.4 & 0.64 \\
PAT3D~\cite{lin2026pat3d}
& 0.046 & 70.8 & 0.68 \\
\midrule
GRPO only
& 0.044
& 70.5
& 0.69 \\

GRPO + SFT
& 0.041
& 73.6
& 0.71 \\

GRPO + SFT + Reward
& \textbf{0.039}
& \textbf{75.5}
& \textbf{0.72} \\
\bottomrule
\end{tabular}
\end{adjustbox}
\end{table}

\noindent\textbf{Geometry reconstruction.}
Besides our physics metrics, we evaluate scene-level reconstruction
quality using Chamfer distance (CD-S), F-score (F-Score-S), and
oriented bounding-box IoU (IoU-B) (\cref{tab:quality_results}).
Compared with MIDI, PhyMix reduces CD-S from 0.072 to 0.039,
increases F-Score-S from 57.2 to 75.5, and improves IoU-B from
0.60 to 0.72.
These improvements indicate that the gains in physical consistency
are accompanied by better agreement with the reference scene
geometry and spatial layout.

\begin{figure*}[t]
\centering
\includegraphics[width=0.99\textwidth,height=0.39\textheight,keepaspectratio]{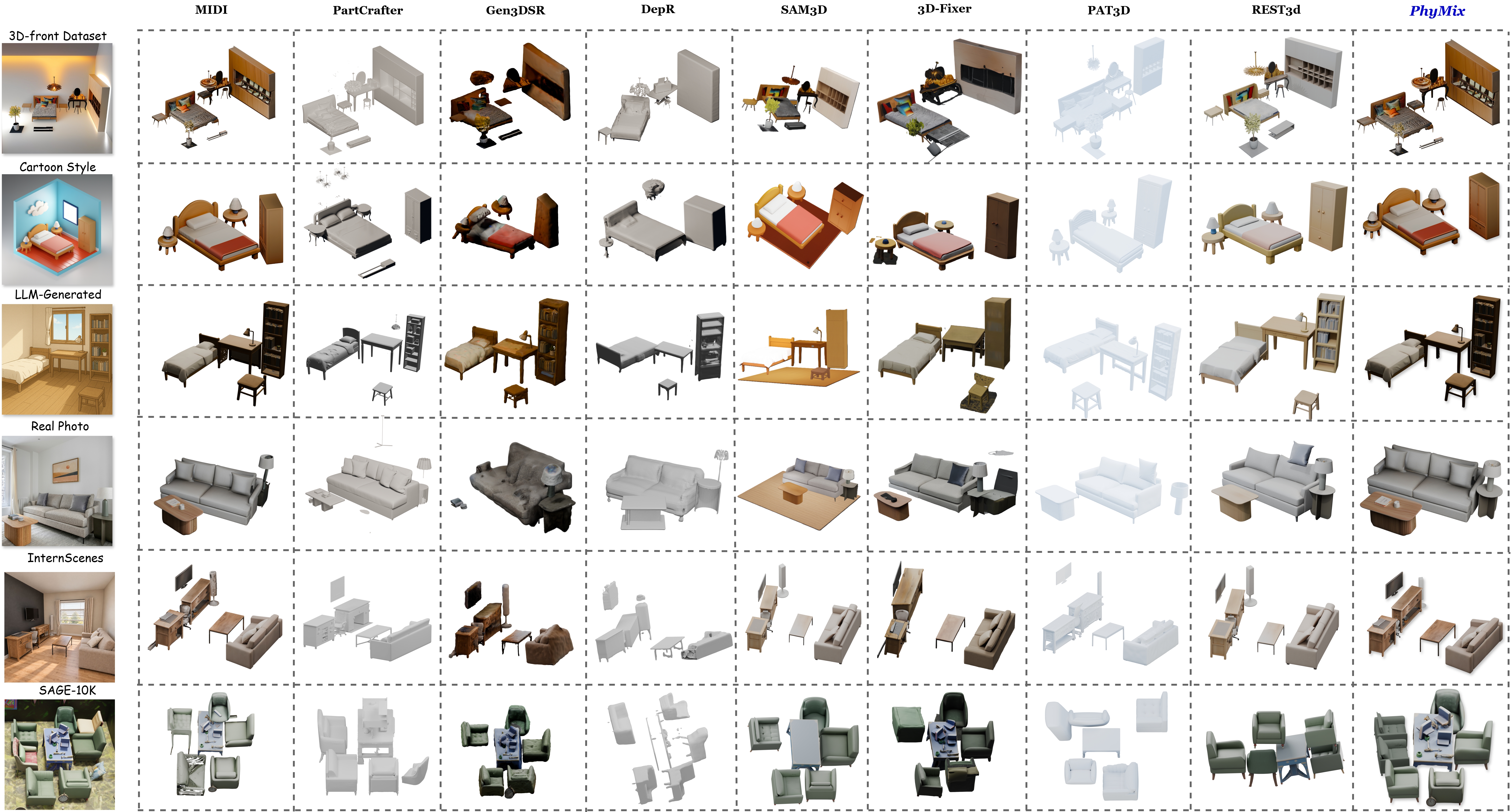}
\caption{Qualitative comparisons on synthetic, stylized, and real-world inputs. Examples illustrate differences in object placement, contact, and geometric reconstruction.}
\label{fig:qual_main}
\end{figure*}

\noindent\textbf{Qualitative comparison.}
\Cref{fig:qual_main} compares results on 3D-FRONT,
InternScenes, and SAGE-10K, together with stylized and
real-world inputs. PhyMix produces coherent object geometry
and preserves the main spatial relationships in the input.
In the InternScenes example, it retains the arrangement
of the sofa, coffee table, and TV cabinet, while several
baselines distort the furniture or separate related components.
The SAGE-10K example contains multiple chairs around a central
table, where some methods alter chair orientations or lose
the seating arrangement. PhyMix better preserves the chairs'
placement around the table and the visible tabletop contents.
These two examples demonstrate reconstruction on datasets
not used for fine-tuning, complementing the quantitative
cross-dataset results. The stylized and real-photo examples
further illustrate its ability to retain recognizable objects
and plausible layouts across varied visual inputs.

\subsection{Ablation Studies}
\label{sec:ablations}
\subsubsection{Backbone and Candidate-Group Settings}
\begin{table*}[!t]
\centering
\caption{Backbone, candidate-group, and reference-regularization comparisons. In (a), MIDI and PartCrafter use GRPO alignment and agentic repair, while PAT3D uses agentic repair only, without generator alignment. Parts (b) and (c) evaluate the full pipeline. Relative time is normalized to the $K=2$ configuration.}
\label{tab:grpo_ablation}
\footnotesize\setlength{\tabcolsep}{3pt}\renewcommand{\arraystretch}{1.0}
\begin{minipage}[t]{0.40\textwidth}\vspace{0pt}\centering
\begin{tabular*}{\linewidth}{@{\extracolsep{\fill}}llcc@{}}
\multicolumn{4}{c}{\textit{(a) Cross-backbone}} \\[1pt]
\toprule
Backbone & Added modules & Base$\uparrow$ & Updated$\uparrow$ \\
\midrule
MIDI & GRPO + Agentic & 85.3 & \textbf{98.6} \\
PartCrafter & GRPO + Agentic & 88.8 & \textbf{96.3} \\
PAT3D & Agentic only & 93.2 & \textbf{97.1} \\
\bottomrule
\end{tabular*}
\end{minipage}
\hfill
\begin{minipage}[t]{0.25\textwidth}\vspace{0pt}\centering
\begin{tabular*}{\linewidth}{@{\extracolsep{\fill}}ccc@{}}
\multicolumn{3}{c}{\textit{(b) Group size $K$}} \\[1pt]
\toprule
$K$ & $\mathrm{Overall}_{\rm rep}\uparrow$ & Rel. time \\
\midrule
2  & 92.3          & $1.0\times$ \\
4  & 95.5          & $1.8\times$ \\
8  & 97.0          & $3.2\times$ \\
12 & \textbf{98.6} & $4.5\times$ \\
16 & 98.5          & $6.1\times$ \\
\bottomrule
\end{tabular*}
\end{minipage}
\hfill
\begin{minipage}[t]{0.31\textwidth}\vspace{0pt}\centering
\begin{tabular*}{\linewidth}{@{\extracolsep{\fill}}lcc@{}}
\multicolumn{3}{c}{\textit{(c) KL regularization}} \\[1pt]
\toprule
Config & $\mathrm{Overall}_{\rm rep}\uparrow$ & CD-S$\downarrow$ \\
\midrule
w/o KL     & 94.8          & 0.052 \\
Fixed KL   & 97.5          & 0.043 \\
Decayed KL & \textbf{98.6} & \textbf{0.039} \\
\bottomrule
\end{tabular*}
\end{minipage}
\end{table*}
\Cref{tab:grpo_ablation}(a) shows that PhyMix improves
the overall physical score from 85.3 to 98.6 with MIDI
and from 88.8 to 96.3 with PartCrafter.
Applying agentic repair alone to PAT3D raises its score
from 93.2 to 97.1 without retraining the generator.
These results demonstrate compatibility with different
backbones and the effectiveness of repair as a standalone stage.
\Cref{tab:grpo_ablation}(b) examines the candidate group size.
Performance improves as $K$ increases from 2 to 12.
Increasing $K$ to 16 raises relative runtime from
$4.5\times$ to $6.1\times$ without further improving
the score. We therefore use $K=12$ to balance quality
and computation.
\Cref{tab:grpo_ablation}(c) shows the benefit of
KL regularization. A fixed KL weight improves both
the physical score and scene-level CD compared with
training without KL. Gradually decreasing the weight
gives the best results, reaching a physical score of
98.6 and a CD-S of 0.039.
These comparisons measure the final results after
alignment and repair.

\subsubsection{Repair Strategy and Verification}
\begin{table}[!htb]
\centering
\caption{Repair configurations on 3D-FRONT. Repair Round Success Rate is the percentage of repair rounds that commit an edit. Avg. Repair Rounds is the number of repair rounds per scene, including rounds that commit no edit (\suppsec{app:metrics}). Bold marks the best result among configurations with repair.}
\label{tab:agentic_ablation}
\footnotesize
\setlength{\tabcolsep}{2pt}
\begin{adjustbox}{max width=\linewidth}
\begin{tabular}{lccccc}
\toprule
Method
& $\mathrm{Overall}_{\rm rep}\uparrow$
& \shortstack{F-Score\\S$\uparrow$}
& IoU-B$\uparrow$
& \shortstack{Avg. Repair\\Rounds$\downarrow$}
& \shortstack{Repair Round\\Success Rate (\%)$\uparrow$} \\
\midrule
MIDI
& 85.3 & 57.2 & 0.60 & 0.0 & -- \\
MIDI + Agentic Repair
& 93.6 & 65.5 & 0.66 & 4.2 & 57.0 \\
GRPO only
& 96.8 & 70.5 & 0.69 & 0.0 & -- \\
GRPO + Heuristic
& 97.3 & 72.0 & 0.70 & 3.4 & 54.2 \\
GRPO + Search
& 98.3 & 74.2 & 0.71 & 3.1 & 63.5 \\
GRPO + SFT
& 98.2 & 73.6 & 0.71 & 2.8 & 61.3 \\
GRPO + SFT + Reward
& \textbf{98.6} & \textbf{75.5} & \textbf{0.72}
& \textbf{2.4} & \textbf{66.8} \\
\bottomrule
\end{tabular}
\end{adjustbox}
\end{table}

\begin{table*}[!t]
\centering
\caption{Skill and verifier ablations on 3D-FRONT.
Selected violation rates are shown as percentages;
$\mathrm{Overall}_{\rm rep}$ is computed from all benchmark rates.}
\label{tab:skill_verifier_ablation}
\footnotesize
\setlength{\tabcolsep}{5pt}
\begin{adjustbox}{max width=\textwidth}
\begin{tabular*}{\linewidth}{@{\extracolsep{\fill}}lcccccc@{}}
\toprule
Config
& Collision$\downarrow$
& Floating$\downarrow$
& Unanchored$\downarrow$
& Misori.$\downarrow$
& Unreach.$\downarrow$
& $\mathrm{Overall}_{\rm rep}\uparrow$ \\
\midrule
Full repair
& \textbf{0.56}
& \textbf{0.97}
& \textbf{1.80}
& \textbf{1.72}
& \textbf{1.01}
& \textbf{98.6} \\
w/o \texttt{separate\_collision}
& 2.04 & 1.05 & 1.95 & 1.81 & 1.14 & 97.6 \\
w/o \texttt{snap\_to\_support}
& 0.68 & 5.42 & 4.36 & 1.90 & 1.32 & 96.8 \\
w/o \texttt{align\_to\_wall}
& 0.59 & 1.12 & 3.51 & 1.78 & 1.18 & 97.7 \\
w/o \texttt{rotate}/\texttt{rescale}
& 0.71 & 1.35 & 2.08 & 3.26 & 1.28 & 97.4 \\
w/o reachability-aware adjustment
& 0.62 & 1.08 & 1.94 & 1.85 & 2.84 & 97.8 \\
w/o verifier
& 0.83 & 1.94 & 3.20 & 2.44 & 1.77 & 96.9 \\
\bottomrule
\end{tabular*}
\end{adjustbox}
\end{table*}

\Cref{tab:agentic_ablation} compares different repair
strategies applied to the aligned generator.
Heuristic repair uses rule-based edits,
while Search tries preset parameters for each repair skill.
Both use the verifier to select and accept edits.
In this comparison, rule-based heuristic repair improves
the overall physical score from 96.8 to 97.3, while searching
over predefined repair actions and their parameters raises
it to 98.3.
The learned SFT policy achieves a similar score of 98.2
with fewer repair rounds (2.8 versus 3.1).
Further reward refinement achieves the highest score
of 98.6 and reduces the average number of rounds to 2.4.
Search already approaches the full method in physical quality; the learned policy's main advantage
in this comparison is fewer repair rounds and a higher round success rate (66.8\% versus 63.5\%), rather than a large score gain.
The verifier continues to check each proposed edit
before acceptance.

\Cref{tab:skill_verifier_ablation} examines the contribution
of individual repair skills and verification.
The w/o verifier variant disables accept-or-rollback
checks and keeps proposed edits without verification.
Removing collision separation increases the collision
rate from 0.56\% to 2.04\%, while removing support snapping
raises the floating rate from 0.97\% to 5.42\%.
Removing support snapping also raises the unanchored rate to 4.36\%, exceeding the 3.51\% observed without wall alignment. This cross-metric effect suggests interactions between support and attachment repairs.
Without reachability adjustment, the proportion of
unreachable start--goal pairs increases from 1.01\% to 2.84\%.
Disabling the verifier also worsens
all reported violation rates.
These results show that the repair skills address their
corresponding physical errors, while verification helps
prevent harmful edits from being retained.

\subsubsection{Action-Selector Cost}
\begin{table}[!htb]
\centering
\caption{Action-selection comparison. Repair Round Success Rate counts rounds with a committed edit. Runtime is normalized by the full PhyMix pipeline (100\%). Lower preservation cost is better.}
\label{tab:vlm_policy}
\footnotesize
\setlength{\tabcolsep}{3pt}
\begin{adjustbox}{max width=\linewidth}
\begin{tabular}{lcccc}
\toprule
\shortstack{Action\\selector} & $\mathrm{Overall}_{\rm rep}\uparrow$ & \shortstack{Pres.\\Cost$\downarrow$} & \shortstack{Runtime\\(\%)$\downarrow$} & \shortstack{Repair Round\\Success Rate (\%)$\uparrow$} \\
\midrule
GRPO + Heuristic & 97.3 & 0.061 & \textbf{82.0} & 54.2 \\
VLM direct repair & 97.2 & 0.083 & 162.0 & 43.6 \\
GRPO + VLM + Verifier & 98.0 & 0.052 & 139.0 & 57.8 \\
Distilled SFT policy & 98.2 & 0.044 & 104.0 & 61.3 \\
SFT + reward & \textbf{98.6} & \textbf{0.039} & 100.0 & \textbf{66.8} \\
\bottomrule
\end{tabular}
\end{adjustbox}
\end{table}

\Cref{tab:vlm_policy} compares VLM-based repair with the learned repair policies after generator alignment. Preservation cost measures how much repair changes the original scene layout and masks; lower values mean fewer changes. Adding a verifier to direct VLM repair improves the overall physical score from 97.2 to 98.0 and reduces preservation cost from 0.083 to 0.052, indicating better physical consistency with less change to the original reconstruction. Distilling verified edits into an SFT policy further raises the score to 98.2. Reward refinement achieves the best score of 98.6 and the lowest preservation cost of 0.039. With its runtime normalized to 100\%, VLM-based repair with verification requires 139.0\%, corresponding to a 28.1\% runtime reduction for the learned policy. Heuristic repair is faster at 82.0\% but achieves a lower score of 97.3. These results show that the learned policy improves repair quality while reducing the time required by online VLM proposals.

\subsection{Comparison with Rigid-Body Simulation}

\begin{table}[!htb]
\centering
\begin{minipage}[t]{0.49\linewidth}\vspace{0pt}
\centering
\caption{Evaluator and Taichi physical rates (\%).}
\label{tab:evaluator_validation}
\footnotesize
\setlength{\tabcolsep}{2.5pt}
\renewcommand{\arraystretch}{1.1}
\begin{adjustbox}{max width=\linewidth}
\begin{tabular}{@{}lcc@{}}
\toprule
Metric & Ours & Taichi \\
\midrule
Collision (\%)      & 0.56 & 0.52 \\
Floating (\%)       & 0.97 & 1.12 \\
Static instab. (\%) & 1.27 & 1.45 \\
\bottomrule
\end{tabular}%
\end{adjustbox}
\end{minipage}\hfill
\begin{minipage}[t]{0.47\linewidth}\vspace{0pt}
\centering
\caption{Relative per-scene runtime (PhyMix = 100\%).}
\label{tab:inference_efficiency}
\footnotesize
\setlength{\tabcolsep}{2.5pt}
\renewcommand{\arraystretch}{1.1}
\begin{adjustbox}{max width=\linewidth}
\begin{tabular}{@{}lc@{}}
\toprule
Method & Runtime (\%) \\
\midrule
PAT3D~\cite{lin2026pat3d} & 663.6 \\
PhyMix (Taichi) & 197.6 \\
PhyMix (Ours) & \textbf{100.0} \\
\bottomrule
\end{tabular}%
\end{adjustbox}
\end{minipage}
\end{table}

\Cref{tab:evaluator_validation} compares our evaluator with a rigid-body simulator
implemented in Taichi~\cite{hu2019taichi} on the PhyMix outputs for 3D-FRONT,
where objects settle under gravity with the physical parameters of \suppsec{app:dynamic}.
The two evaluations yield similar collision, floating, and static-instability rates,
with absolute differences of only 0.04, 0.15, and 0.18 percentage points, respectively.
This agreement supports the three geometric tests of our evaluator on near-valid scenes. 
\Cref{tab:inference_efficiency} reports complete-pipeline runtimes normalized by the full PhyMix pipeline (100\%).
Replacing our three geometric tests with Taichi simulation raises the runtime to 197.6\%.
For context, the per-scene runtime reported in the original PAT3D paper~\cite{lin2026pat3d} corresponds to 663.6\%.
Besides these three tests, our evaluator also checks object orientation, scale,
required wall or ceiling attachments, and path accessibility,
providing more complete feedback for scene alignment and repair.

\begin{figure}[!t]
\centering
\includegraphics[width=\linewidth]{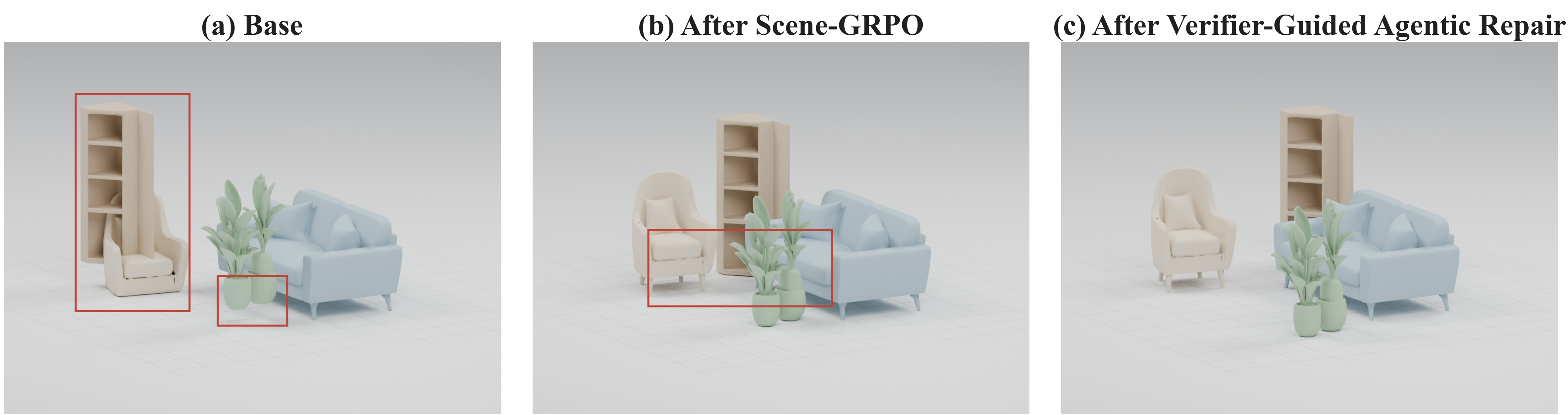}
\caption{Stage-wise comparison: (a) base, (b) Scene-GRPO,
and (c) verifier-guided agentic repair.
Red boxes highlight problematic regions.}
\label{fig:stage_comparison}
\end{figure}

\subsection{Stage-Wise Qualitative Analysis}

\Cref{fig:stage_comparison} illustrates the complementary effects of generator alignment and agentic repair.
The base output contains overlapping object placements.
Scene-GRPO reduces collisions, but reachability violations remain.
Verifier-guided agentic repair further adjusts object placements to address these residual reachability violations.

\subsection{Evidence Beyond Our Physics Metrics}
\label{sec:evaluation_validity}

The preceding experiments show that PhyMix improves scene quality beyond the physics scores used for training and repair. Human ratings and paired comparisons support improved perceived plausibility (\cref{tab:phys_benchmark}), while better Chamfer distance, F-score, and bounding-box IoU demonstrate more accurate reconstruction (\cref{tab:quality_results}). These improvements are measured through human judgments and agreement with reference geometry, rather than through our physics scores alone.  The Taichi comparison further supports our evaluator: collision, floating, and static-instability rates differ by only 0.04, 0.15, and 0.18 percentage points, respectively (\cref{tab:evaluator_validation}).

All methods are evaluated under the same rules and thresholds, ensuring consistent comparison. The gains on held-out scenes and on InternScenes and SAGE-10K without additional fine-tuning also demonstrate that the improvements extend beyond the training examples. Together, these findings support that PhyMix learns to produce better scenes, rather than merely achieving higher scores under its optimization criteria.

\subsection{Additional Qualitative Analysis}
Extended visual examples are provided in \suppsec{Additional Experiment}, and failure cases in \suppsec{app:failure_cases}. These illustrate limitations such as deformable objects, segmentation errors, and uncertain support relations. 

\section{Conclusion}
We presented PhyMix, a framework that connects physical evaluation, generator alignment, and verifier-guided scene repair. The Physics Evaluator organizes nine constraints into geometric priors, contact, stability, and deployability, providing both scene-level feedback and object-level diagnostics. Scene-GRPO uses this feedback to guide generation preferences, while agentic repair applies bounded, verified edits to residual violations in individual outputs. Together, the two stages connect learned physical preferences with scene-specific correction under content-preservation constraints.

Experiments show improvements in physical plausibility and reconstruction quality, and component analyses illustrate the roles of alignment, repair skills, and verification. Human ratings and qualitative examples further reflect improvements in perceived scene plausibility. More broadly, PhyMix treats physical consistency as a measurable goal of scene generation rather than a by-product of visual fidelity, using one evaluator to guide both training and repair. Because it builds on existing generators without changing their architecture, it can be applied to other object-level scene generators. We hope this work helps move single-image scene generation toward scenes that are ready for simulation, robotics, and embodied AI. Remaining limitations are discussed in \suppsec{app:limitations}.

\clearpage
\section*{Supplementary Material}
\noindent This supplement provides details of the physics evaluator, alignment objective, experimental settings, repair protocol, and additional experiments.
\appendix
\section{Unified Physics Evaluator}
\label{sec:subevaluator}
This appendix defines the nine physical tests used in the main paper. \Cref{app:eval_scope} explains how the evaluator decides which objects each test checks, \cref{app:eval_tests} defines the tests, \cref{app:feedback_aggregation} explains how test results become the reported scores, and \cref{app:repair_cost} gives the cost that guides repair.

\subsection{What Each Object Is Checked Against}
\label{app:eval_scope}
\noindent\textbf{What each object should rest on.}
To check contact, the evaluator must know what each object is supposed to rest on. A generated scene cannot tell us this: a lamp floating in the air could belong on the table or on the floor. We therefore decide it from the input image before evaluating any method. Each object is assigned to the floor, another object, a wall, or the ceiling, and every method is checked against the same assignment. For example, if the lamp in the image stands on a table, a generated lamp floating above the table fails. The room's floor and walls and each object's category are also estimated from the input and shared by all methods.

\medskip
\noindent\textbf{Which tests an object receives.}
The assignment decides the contact test. An object on the floor is checked by \emph{grounding}, an object on another object by \emph{support}, and an object on a wall or ceiling by \emph{anchoring}. Sometimes the image does not show what an object rests on, for example when it is heavily occluded. If such an object usually stands on the floor, like a chair or a bed, we assign it to the floor. Otherwise, we do not know what it should touch, so it skips the contact and static-stability tests; it is still checked for orientation, scale, collision, and dynamic stability.

\medskip
\noindent\textbf{Distance thresholds.}
Thresholds on gaps and penetrations scale with the scene, so that the same values work for small and large rooms. They are multiples of
\begin{equation}
 L=\tfrac12D_{\rm room}+\tfrac12\bar d,
 \label{eq:reference_length}
\end{equation}
where $D_{\rm room}$ is the diagonal of the room and $\bar d$ is the mean diagonal of the objects, both estimated from the input. For a room with a 7 m diagonal and objects with a 1 m mean diagonal, $L=4$ m, so the contact threshold $0.002L$ is 8 mm. All lengths are in meters, and the vertical axis $e_z$ points up.

\subsection{Test Definitions}
\label{app:eval_tests}
\Cref{tab:nine_constraints} summarizes the nine tests; each is defined below.

\begin{table*}[t]
\centering
\caption{The nine physical tests. A violation rate is the fraction of checked objects (or pairs) that fail. $L$ is defined in \cref{eq:reference_length}.}
\label{tab:nine_constraints}
\small\renewcommand{\arraystretch}{1.12}
\begin{tabularx}{\textwidth}{@{}p{1.9cm}p{3.6cm}X@{}}
\toprule
Test & Checked objects & Fails if \\
\midrule
Orientation & Objects whose category has an upright direction & tilt exceeds $\theta_o=15^\circ$ \\
Scale & Objects whose category has $\ge$20 training examples & a dimension lies outside $[0.8Q_{0.05},\,1.2Q_{0.95}]$ for the category \\
Collision & All object pairs and object--room pairs & overlap ratio $>\eta_c=0.005$ \emph{and} penetration $>\tau_c=0.001L$ \\
Grounding & Objects assigned to the floor & gap $>\tau_g=0.002L$, penetration $>\tau_c$, or contact area $<1\%$ of the footprint \\
Support & Objects assigned to another object & the supporting object is missing, or the gap, penetration, or contact-area check fails \\
Anchoring & Objects assigned to a wall or ceiling & mounting gap $>\tau_a=0.002L$ or mounting angle error $>\theta_a=10^\circ$ \\
Static stability & Objects assigned to the floor, another object, a wall, or the ceiling & center of mass falls more than $\tau_s=0.002L$ outside the contact region, or the mount fails anchoring \\
Dynamic stability & All movable objects & after a small push, it moves $>0.02L$, rotates $>10^\circ$, or keeps moving \\
Reachability & 100 fixed start--goal pairs & an endpoint is blocked, or no path connects the pair \\
\bottomrule
\end{tabularx}
\end{table*}

\medskip
\noindent\textbf{Orientation.}
This test detects objects that should stand upright but are tilted, such as a chair lying on its side. Each category has an upward axis, and the tilt is the angle between this axis and $e_z$. Only tilt is checked; turning an object around the vertical axis is allowed.

\medskip
\noindent\textbf{Scale.}
This test detects objects with implausible sizes, such as a bed smaller than a chair. For each category, we take the 5th and 95th percentiles, $Q_{0.05}$ and $Q_{0.95}$, of width, depth, and height in the training set, and widen this range by 20\% at each end. An object fails if any dimension lies outside the range. Categories with fewer than 20 training examples are skipped, because their percentiles are unreliable.

\medskip
\noindent\textbf{Collision.}
This test detects objects that pass through each other. For two objects, we measure the \emph{overlap ratio}, their shared volume divided by the volume of the smaller object, and the \emph{penetration}, the deepest point of one object inside the other. A pair collides only if both exceed their thresholds, so a book lying on a table, which touches it with a tiny numerical overlap, does not count. An object that extends through a wall or the floor is detected in the same way, using the part of the object outside the room. Volumes and depths are computed on a voxel grid with spacing $0.001L$; meshes with holes are first replaced by closed versions. The collision rate is the fraction of colliding pairs among all $N(N-1)/2$ object pairs and $N$ object--room pairs, where $N$ is the number of objects.

\medskip
\noindent\textbf{Grounding and support.}
These tests detect objects that do not rest properly on their assigned surface: the floor for grounding, and the supporting object for support. An object fails in three cases:
\begin{itemize}
 \item it \emph{floats}: the gap between its bottom and the surface exceeds $\tau_g$;
 \item it \emph{sinks in}: it penetrates the surface by more than $\tau_c$;
 \item it barely touches: less than 1\% of its footprint lies within $0.002L$ of the surface, as for a sofa tilted so that only one corner touches the floor.
\end{itemize}
Support also fails if the supporting object is missing from the output.

\medskip
\noindent\textbf{Anchoring.}
This test detects wall- or ceiling-mounted objects, such as wall shelves and pendant lamps, that are not attached properly. For a wall mount, we measure the gap between the object's back and the wall, and the angle between the object's back and the wall. For a ceiling mount, we measure the gap at the mounting point and the angle between the hanging direction and the vertical. Only objects assigned to a wall or ceiling are checked, so a wardrobe standing near a wall need not touch it.

\medskip
\noindent\textbf{Static stability.}
This test detects objects that touch their support but would tip over. We compute the object's center of mass, assuming uniform density, and check that it lies above the region where the object touches its support, within $\tau_s$. An object that touches nothing fails. We apply the same check to a support together with everything on it, so a table loaded heavily on one edge can fail even if the empty table is stable. A mounted object has no contact region; it passes static stability exactly when it passes anchoring, so a failed mount is counted in both tests.

\medskip
\noindent\textbf{Dynamic stability.}
\label{app:dynamic}
This test detects objects that are balanced but would move after a small disturbance, such as a lamp that falls when bumped. We run four two-second simulations in MuJoCo. In each, every movable object starts with a speed of $0.05\,\mathrm{m/s}$ in one horizontal direction: $+x$, $-x$, $+y$, or $-y$. An object fails if, in any simulation, it moves more than $0.02L$ or turns more than $10^\circ$, or if at the end it is still moving faster than $0.01L$ per second or $5^\circ$ per second. Each object is simulated as one rigid body, and the room does not move. A mounted object is fixed to its wall or ceiling only if it passes anchoring; otherwise it can fall. Settings: gravity $9.81\,\mathrm{m/s^2}$, time step $1/240$ s, 50 solver iterations, friction 0.5, density $500\,\mathrm{kg/m^3}$, and MuJoCo's default contact parameters.

\medskip
\noindent\textbf{Reachability.}
This test checks whether a robot can move around the room. We draw a floor map with 0.02 m cells and mark as blocked every cell under an object part lower than 1.5 m, plus the room boundary. Blocked areas are widened by the robot radius of 0.25 m. We then pick 100 start--goal pairs, with a fixed random seed, from the free floor of the input scene, and search for a path between each pair with A*. A pair fails if either point is blocked or no path exists. The same 100 pairs are used for every method and throughout repair.

\subsection{Scores}
\label{app:feedback_aggregation}
Each test gives a violation rate per scene. These rates are combined in two ways: $Q_9$ for training and $\mathrm{Overall}_{\rm rep}$ for the tables in the main paper.

\medskip
\noindent\textbf{Violation rate.}
For test $j$, let $\mathcal E_j$ be the checked objects (or pairs) in scene $S$, and let $v_{je}=1$ if $e$ fails and $0$ otherwise. The violation rate is the fraction that fails:
\begin{equation}
 P_j(S)=\frac{1}{|\mathcal E_j|}\sum_{e\in\mathcal E_j}v_{je}.
\end{equation}
If a scene has nothing to check for a test, such as anchoring in a room without mounted objects, that test is left out for that scene instead of being counted as 0\%. The dataset-level rate of a test is the mean of its scene rates.

\medskip
\noindent\textbf{Training score $Q_9$.}
Scene-GRPO averages the rates of the tests that apply to the scene, $\mathcal J(S)$:
\begin{equation}
 P(S)=\frac{1}{|\mathcal J(S)|}\sum_{j\in\mathcal J(S)}P_j(S),
 \qquad Q_9(S)=100\bigl(1-P(S)\bigr).
\end{equation}

\medskip
\noindent\textbf{Reported score $\mathrm{Overall}_{\rm rep}$.}
The tables in the main paper combine grounding and support into one \emph{Floating} column, because both detect objects that do not rest on their surface. Floating is the fraction of objects assigned to the floor or to another object that fail their contact test. The tables therefore have eight columns, and $\mathrm{Overall}_{\rm rep}$ is 100 minus their mean (main paper, Eq.~(18)).

\medskip
\noindent\textbf{Severity.}
Violation rates show how often errors occur, not how large they are. Two further measures show size, both multiplied by 100:
\begin{itemize}
 \item \emph{Collision Severity}: the mean overlap ratio, capped at 1, over all pairs; pairs that do not overlap count as zero.
 \item \emph{Floating Gap Severity}: over floating objects only, the mean of $\min\bigl(1,(g-\tau_g)/(0.02L)\bigr)$, where $g$ is the gap. It is 100 when every floating object is at least $0.02L$ too far from its surface, and 0 when nothing floats.
\end{itemize}

\subsection{Graded Cost for Repair}
\label{app:repair_cost}
A test result is only pass or fail, so it cannot show partial progress: reducing a tilt from $30^\circ$ to $20^\circ$ is an improvement, but the object still fails. Repair therefore uses a cost that decreases as an error shrinks.

\medskip
\noindent\textbf{Normalized error.}
For each checked object or pair, the normalized error $b_{je}$ is the measured error divided by its threshold, so $b_{je}>1$ means failure. When a test measures several quantities, the worst one is used:
\begin{itemize}
 \item orientation: tilt$/\theta_o$;
 \item scale: how far the worst dimension is outside its range, as lower bound$/$size or size$/$upper bound;
 \item grounding and support: the largest of gap$/\tau_g$, penetration$/\tau_c$, and required$/$observed contact area;
 \item anchoring: the larger of gap$/\tau_a$ and angle error$/\theta_a$;
 \item static stability: the distance of the center of mass outside the contact region, divided by $\tau_s$;
 \item dynamic stability: the largest ratio of a measured motion or final speed to its limit;
 \item collision: the \emph{smaller} of overlap ratio$/\eta_c$ and penetration$/\tau_c$, because a pair stops colliding as soon as either falls below its threshold.
\end{itemize}

\medskip
\noindent\textbf{Cost.}
The cost of one object or pair is
\begin{equation}
 c_{je}=\frac{\max(0,b_{je}-1)}{1+\max(0,b_{je}-1)}.
 \label{eq:entity_repair_cost}
\end{equation}
It is 0 when the error is within the threshold, increases as the error grows, and never reaches 1, so one very large error cannot dominate the total. A missing support, zero contact area, and a failed reachability pair have cost 1. The scene cost is averaged in the same way as $P(S)$:
\begin{equation}
 \mathcal C(S)=\frac{1}{|\mathcal J(S)|}
 \sum_{j\in\mathcal J(S)}\frac{1}{|\mathcal E_j|}
 \sum_{e\in\mathcal E_j}c_{je}(S).
 \label{eq:continuous_violation}
\end{equation}
$\mathcal C(S)$ is 0 exactly when all tests pass. \Cref{app:repair_utility} describes how repair uses it.

\section{Candidate Scoring in Scene-GRPO}
\label{app:ref_lik}
Scene-GRPO needs a score for each candidate scene that the generator can raise or lower through training. A flow-matching generator has no tractable likelihood, so we use its flow-matching loss on the candidate as this score. This appendix defines the loss and the schedule of the KL weight.

\medskip
\noindent\textbf{Encoding a candidate.}
Each candidate is scored in the same form that the evaluator checks. Any pose or scale perturbation is applied first; the scene is then encoded into the backbone's training representation $x^{(k)}$, with a fixed object order and the training-set normalization.

\medskip
\noindent\textbf{Flow-matching loss.}
We add Gaussian noise at time $t$ and ask the generator to predict the velocity from the clean candidate to the noise:
\begin{equation}
 x_t^{(k)}=(1-t)x^{(k)}+t\epsilon,\qquad
 u_t^{(k)}=\epsilon-x^{(k)},\qquad\epsilon\sim\mathcal N(0,I).
 \label{eq:flow_path}
\end{equation}
The loss for one draw is the mean squared error over the $d_x$ scored coordinates,
\[
 \mathcal L_{\rm FM}
 =\frac{1}{d_x}\bigl\|v_\theta(x_t^{(k)},t,X)-u_t^{(k)}\bigr\|_2^2 ,
\]
and the candidate loss $\ell_\theta^{(k)}$ is its mean over four draws of $(t,\epsilon)$, with $t\in[0.01,0.99]$. All candidates in a group use the same four draws, so their losses differ only because the candidates differ. If a backbone predicts noise instead of velocity, its squared error equals $(1-t)^2$ times the velocity error, so we divide it by $(1-t)^2$.

\medskip
\noindent\textbf{KL weight.}
The KL weight is large early in training and decreases with a cosine schedule. At update $u$ of $U$,
\begin{equation}
 \beta_u=\beta_{\min}+\tfrac12(\beta_{\max}-\beta_{\min})
 \bigl[1+\cos\bigl(\pi\min(u/U,1)\bigr)\bigr],
 \label{eq:kl_schedule}
\end{equation}
with $\beta_{\max}=0.1$ and $\beta_{\min}=0.01$.

\section{User Study}
\label{US}
Thirty participants viewed renderings of each method's output, produced with the same camera and lighting. Method names were hidden, and scene order and left/right positions were randomized.

\medskip
\noindent\textbf{Ratings.}
Participants rated each scene from 1 (worst) to 7 (best) on three attributes: \emph{geometric prior} (objects are upright and plausibly sized), \emph{contact} (objects rest on their supports without floating or interpenetrating), and \emph{stability} (objects look as if they would stay in place). Ratings are averaged per scene and then over scenes. Overall MOS is the mean of the three attribute scores.

\medskip
\noindent\textbf{Paired comparisons.}
Participants saw two outputs for the same input and chose the more physically plausible one, or declared a tie. Each pair of methods was compared on 200 scenes, and a tie counts as half a win.

\section{Experimental Settings}
\label{set}
This appendix describes how the experiments were run. \Cref{app:baselines} gives the data and the inputs of each method, \cref{app:impl} the training settings, \cref{app:skills} the edits that repair can make, \cref{app:repair_utility} how repair decides whether to keep an edit, and \cref{app:metrics} the metrics.

\subsection{Data and Inputs}
\label{app:baselines}
\noindent\textbf{Datasets.}
We train and test on 3D-FRONT, using the MIDI preprocessing pipeline~\cite{fu20213d,huang2025midi}: 12K scenes for training and 4.8K for testing. The two sets are split by room, so no room appears in both. We keep 10\% of the training rooms as a validation set, which we use to choose thresholds and temperatures. InternScenes~\cite{zhong2025internscenes} and SAGE-10K~\cite{xia2026sage} are used only for testing, to check whether PhyMix works on data it was not trained on.

\medskip
\noindent\textbf{Inputs of each method.}
The compared methods have different native input requirements.
For a controlled comparison, all methods are provided with the same RGB image, instance masks, and camera parameters.
Synthetic masks come from the dataset; real and stylized
images use shared segmentation results.
We adapt PAT3D~\cite{lin2026pat3d} to accept image inputs. We skip its reference-image generation stage and use the test image directly for reconstruction and physics optimization. We also supply the shared instance masks and camera parameters through its input interfaces.
Ground-truth 3D scenes are used only for evaluation. All methods use the same test images and random seeds.
Failed generations count as failures on every test rather than being excluded.

\subsection{Training Settings}
\label{app:impl}
PhyMix trains two components: the generator, aligned with Scene-GRPO, and the repair policy, which proposes edits. \Cref{tab:implementation_settings} lists their settings. MIDI and PartCrafter use the same generator settings.

\begin{table*}[t]
\centering
\caption{Training settings of the generator (top) and the repair policy (bottom). EMA: exponential moving average.}
\label{tab:implementation_settings}
\small\renewcommand{\arraystretch}{1.12}
\begin{tabularx}{\textwidth}{@{}p{3.3cm}X@{}}
\toprule
Component & Settings \\
\midrule
Optimizer & AdamW; learning rate $10^{-5}$; betas $(0.9,0.999)$; weight decay 0.01; gradient clipping 1.0 \\
Training & 10,000 updates; batch 4 with 8 accumulation steps (effective batch 32) \\
LoRA & Rank 16, scale 32, dropout 0.05, on layout-conditioning attention and pose heads; geometry encoder and decoder frozen \\
Candidates per input & 12: six from the current model and six perturbed samples from the EMA model (decay 0.999) \\
Perturbations & Translation $\le0.02L$, rotation $\le10^\circ$, log-scale change $\le0.05$ \\
Objective & $\tau_p=\tau_r=1$; advantage clip $a=3$; $\lambda_{\rm align}=\lambda_{\rm FM}=1$; KL weight from 0.1 to 0.01 (\cref{eq:kl_schedule}) \\
Generation & 50 sampling steps; classifier-free guidance 3.0; seeds 0, 1, 2 \\
\midrule
Data collection & GPT-5.6-Sol proposes edits; 10,000 repair trajectories from training rooms; temperature 0.7 \\
Policy training & Qwen3.5-9B; AdamW; learning rate $2\times10^{-5}$; batch 64; three epochs; LoRA rank 16, scale 32; loss terms in main paper Eq.~(14) weighted equally \\
Reward refinement & One epoch; $\mu=\nu=0.1$; $g_0=0.01$, $w_{\max}=4$ \\
\bottomrule
\end{tabularx}
\end{table*}

\medskip
\noindent\textbf{Repair policy.}
We first run repair on training rooms with GPT-5.6-Sol proposing the edits, and keep the edits that the verifier accepts. We then train Qwen3.5-9B on these examples to propose edits by itself.
\begin{itemize}
 \item \emph{What the policy sees:} the input image and masks, a rendering of the current scene, the position of every object, and the evaluator's report of which tests fail, for which objects, and by how much.
 \item \emph{What it outputs:} one edit, such as ``move the lamp down by 2 cm'', or STOP. An edit consists of an action from \cref{tab:skills}, the object to edit, and the amount.
 \item \emph{How it is trained:} it learns to reproduce the accepted edits. It also learns to predict whether an edit will be accepted; this prediction is the score $s_\psi(o,a)$ in the ranking loss (main paper, Eq.~(15)).
\end{itemize}
Each repair round tests up to eight edits. The policy proposes one, and the remaining slots are filled with simple rule-based edits.

\medskip
\noindent\textbf{Reward weighting.}
In the final training stage, edits that remove more error receive more weight. An edit that reduces the cost $\mathcal C$ by $g$ receives weight $w(g)=1+\min(g/g_0,\,w_{\max})$. With the values in \cref{tab:implementation_settings}, a very small improvement has weight 1, and an improvement of 0.04 or more has the maximum weight of 5.

\subsection{What Repair Can Change}
\label{app:skills}
Repair only moves, rotates, and rescales objects. It never adds, removes, or reshapes an object. \Cref{tab:skills} lists the seven available edits. Each has a limit per edit and a total limit relative to the initial scene $S_0$, so that the repaired scene stays close to the input image. When an object is moved, everything standing on it moves with it; for example, moving a table also moves the lamp on it.

\begin{table*}[t]
\centering
\caption{The seven repair edits. Totals are measured from the initial scene $S_0$. The limits of the first three rows apply to every edit.}
\label{tab:skills}
\small\renewcommand{\arraystretch}{1.12}
\begin{tabularx}{\textwidth}{@{}p{3.3cm}X@{}}
\toprule
Edit & What it does \\
\midrule
Translate & Moves an object by at most $0.02L$ per edit and $0.1L$ in total \\
Rotate & Rotates an object about its contact with its support, by at most $15^\circ$ per edit and $30^\circ$ in total \\
Rescale & Changes an object's log-scale by at most 0.05 per edit (about 5\%) and 0.15 in total \\
Separate collision & Pushes two overlapping objects apart along the shortest direction \\
Snap to support & Moves an object straight down or up onto its support surface \\
Align to wall & Moves a wall- or ceiling-mounted object back onto its mounting surface \\
Reachability adjustment & Moves an object that blocks a path toward nearby free space \\
\bottomrule
\end{tabularx}
\end{table*}

\subsection{Deciding Whether to Keep an Edit}
\label{app:repair_utility}
Each proposed edit is first applied to a copy of the scene. The verifier then decides whether to keep it. A good edit should (1) reduce physical errors, (2) keep the scene close to the input image, and (3) be small. For example, suppose a chair floats 3 cm above the floor:
\begin{itemize}
 \item moving it down by 3 cm removes the error with a small edit, so it is \emph{kept};
 \item moving it 0.5 m away to another spot exceeds the movement limit, so it is \emph{rejected};
 \item moving it down but into a table fixes the floating but creates a severe collision, so it is also \emph{rejected}.
\end{itemize}
The three criteria are measured as follows.

\medskip
\noindent\textbf{(1) Physical error}
is the cost $\mathcal C(S)$ from \cref{app:repair_cost}. It decreases as errors shrink.

\medskip
\noindent\textbf{(2) Distance from the input}
has two parts. The layout part measures how far objects have moved, turned, and changed size in scene $S$ compared with a reference scene $R$:
\begin{equation}
\begin{split}
 D_{\rm lay}(S,R)=\frac1N\sum_i\bigg[
 &\frac{\|t_i-t_i^R\|_2^2}{L^2}
 +\frac{d_R(q_i,q_i^R)^2}{\pi^2}\\
 &+\|\log(s_i/s_i^R)\|_2^2\bigg],
\end{split}
\label{eq:layout_distance}
\end{equation}
where $t_i$, $q_i$, and $s_i$ are the position, rotation, and scale of object $i$, $d_R$ is the angle between two rotations, and $N$ is the number of objects. Repair uses the initial scene as the reference, $R=S_0$; Scene-GRPO uses the same measure with each candidate's reference scene (main paper, Eq.~(6)). The image part, $D_{\rm mask}$, measures how well the rendered objects still match the input masks, as the mean of $1-\mathrm{IoU}$ over visible objects. Together,
\begin{equation}
 D(S,S_0;X)=D_{\rm lay}(S,S_0)+D_{\rm mask}(S;X).
 \label{eq:preservation}
\end{equation}
Every scene is compared with the initial scene $S_0$, not with the previous one. Otherwise, many small edits could add up to a large change.

\medskip
\noindent\textbf{(3) Edit size}
counts how much the edit moves, turns, and rescales each affected object, plus a fixed cost of 1 for making any edit at all:
\begin{equation}
 c(a;N)=\frac1N\bigg[1+\sum_{i\in\mathcal A(a)}\Big(
 \frac{\|\Delta t_i\|}{L}+\frac{\Delta\phi_i}{\pi}
 +\|\Delta\log s_i\|_1\Big)\bigg],
\end{equation}
where $\mathcal A(a)$ contains the edited object and everything that moves with it.

\medskip
\noindent\textbf{Decision.}
The three criteria are combined into one gain:
\begin{equation}
\begin{aligned}
\Delta J
&=\underbrace{\bigl[\mathcal C(S_\ell)-\mathcal C(\tilde S)\bigr]}_{\text{less error}} \quad-\underbrace{\bigl[D(\tilde S,S_0;X)-D(S_\ell,S_0;X)\bigr]}_{\text{further from input}}\\
&\quad-\underbrace{10^{-3}\,c(a;N)}_{\text{edit size}}.
\end{aligned}
\end{equation}
where $S_\ell$ is the current scene and $\tilde S$ the edited one. This is the gain in main paper Eq.~(12) with $\lambda_{\rm pres}=\lambda_{\rm img}=1$ and $\lambda_{\rm cost}=10^{-3}$. An edit is kept only if all of the following hold:
\begin{enumerate}
 \item it reduces the physical error $\mathcal C$;
 \item it stays within the limits in \cref{tab:skills}, and the scene stays close to the input: $D_{\rm lay}(\tilde S,S_0)\le0.01$, and the mask error rises by at most 0.02;
 \item it creates no severe error: no object's normalized error $b_{je}$ may rise above 2 or grow if it is already above 2, and no more than ten reachability pairs may become blocked;
 \item it has the largest gain among the proposed edits, and this gain exceeds $10^{-4}$.
\end{enumerate}
Condition 3 is needed because the average error can fall even when one object gets much worse, as in the chair example above.

\medskip
\noindent\textbf{When repair stops.}
If none of the proposed edits is kept, repair tries one more set of up to eight rule-based edits on the largest remaining error. Repair stops when all tests pass, when these rule-based edits also fail, or when a budget runs out: 8 kept edits, 16 rounds, or 256 tested edits.

\subsection{Metrics and Runtime}
\label{app:metrics}
\noindent\textbf{Geometry metrics.}
These metrics compare the generated scene with the ground-truth scene. We sample 100,000 points on the surfaces of each scene. \emph{CD-S} (Chamfer distance) is the mean squared distance between the two point sets, in $\mathrm{m}^2$; lower is better. \emph{F-Score-S} is the percentage of points that lie within 2 cm of the other scene, combining precision and recall; higher is better. \emph{IoU-B} is the mean overlap between the generated and ground-truth bounding box of each object; higher is better. Objects are matched by instance ID, and a missing object counts as unmatched. All three metrics are computed in the same world frame without extra alignment, so an object in the wrong place lowers the score even if its shape is correct.

\medskip
\noindent\textbf{Repair statistics.}
A repair round tests a set of edits and keeps at most one. \emph{Repair Round Success Rate} is the percentage of rounds that keep an edit. \emph{Avg.\ Repair Rounds} is the average number of rounds per scene; a scene that needs no repair counts as zero.

\medskip
\noindent\textbf{Runtime.} Runtime is the total time from the input image to the final scene,
including generation, repair, and API calls; offline data collection and training are excluded.
Because absolute times depend on hardware and on the latency of online API calls,
we report runtimes relative to the full PhyMix pipeline, measured on the same machine.
\section{Additional Results}
\label{Additional Experiment}

\begin{figure*}[t]
    \centering
    \includegraphics[width=\linewidth]{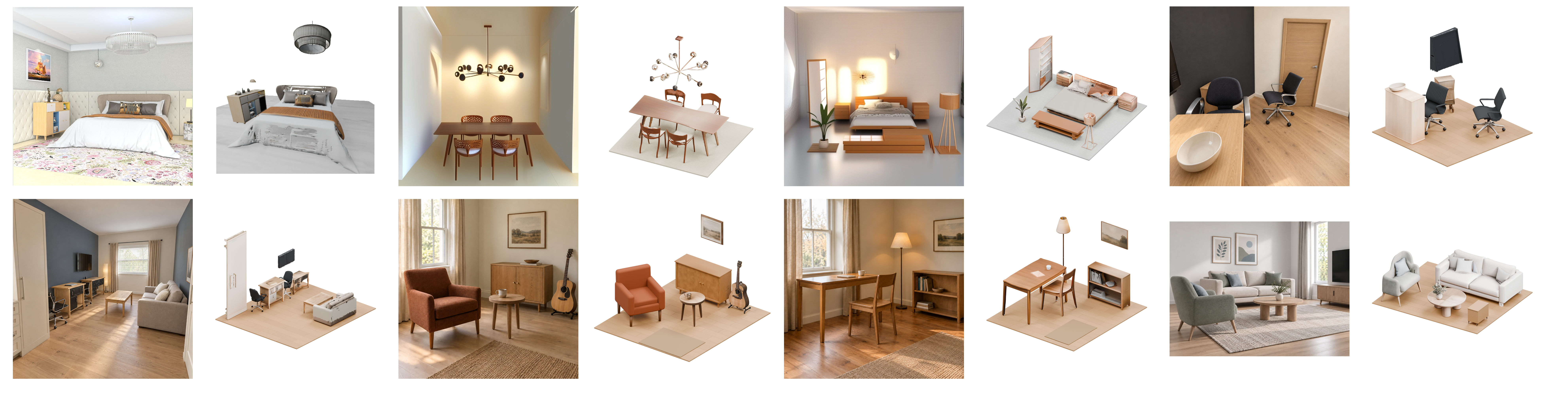}
    \caption{Additional PhyMix results across diverse indoor scenes and visual styles. Each pair shows an input image on the left and the reconstructed 3D scene on the right.}
    \label{fig:extended_comparison}
\end{figure*}

\Cref{fig:extended_comparison} shows additional PhyMix results on a range of indoor scenes, including bedrooms, living rooms, dining rooms, and workspaces. Each input image is paired with its reconstructed 3D scene. These examples illustrate how PhyMix captures the objects and their spatial arrangement across different room layouts and visual styles. 

\section{Failure Cases and Limitations}
\label{app:limitations}
This appendix shows typical failure cases (\cref{app:failure_cases}) and then summarizes the limitations of PhyMix (\cref{app:limits}).

\subsection{Failure Cases}
\label{app:failure_cases}
\begin{inlinefigure}
\centering
\includegraphics[width=\columnwidth,height=0.32\textheight,keepaspectratio]{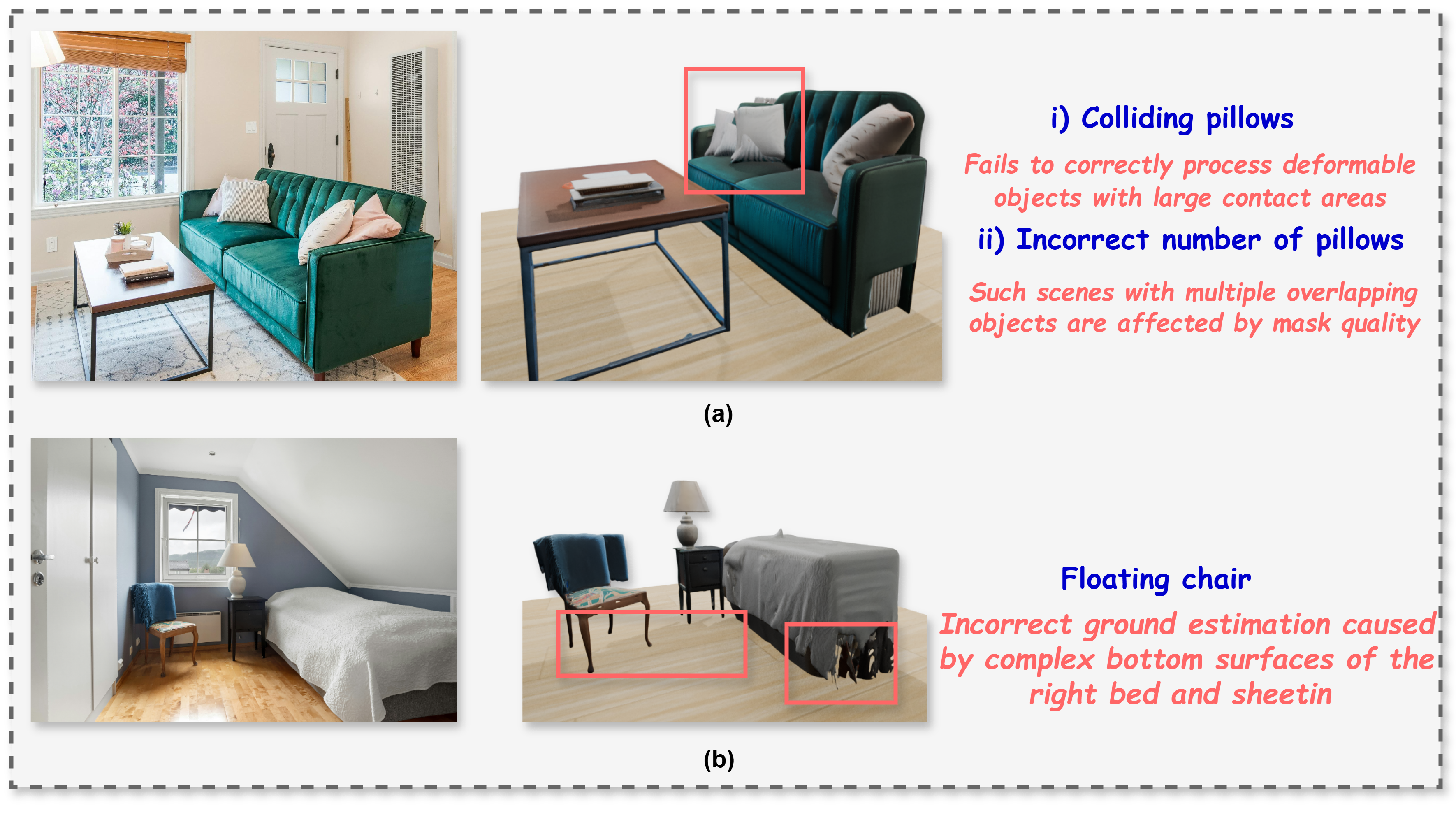}
\caption{Failure cases. (a) The pillows collide with the sofa because rigid meshes cannot deform, and overlapping pillows merged in the masks give the wrong number of pillows. (b) The bed sheet hides the bottom of the bed, the floor is estimated incorrectly, and the chair floats.}
\label{fig:failure_cases}
\end{inlinefigure}
\Cref{fig:failure_cases} shows three failures that PhyMix does not fix. In all three, the cause lies in the object model or the input estimates rather than in the repair search.

\medskip
\noindent\textbf{Colliding pillows (a).}
A real pillow deforms where it presses into a sofa, so it touches the sofa over a large area. Our meshes are rigid. A pillow placed in full contact with the sofa passes into it and counts as a collision, while a pillow moved out of the sofa loses contact and floats. No placement satisfies both tests, so the collision remains.

\medskip
\noindent\textbf{Wrong number of pillows (a).}
Overlapping pillows are hard to separate in the input masks. When two pillows share one mask, the generator produces one object in their place. Repair keeps the set of objects fixed, so it cannot restore the missing pillow.

\medskip
\noindent\textbf{Floating chair (b).}
The floor is estimated from the input image. Here the bed sheet hangs down to the floor and hides the bottom of the bed, which makes the floor estimate wrong. Grounding is then checked against this wrong floor, so the evaluator does not report the chair as floating and repair does not move it.
\subsection{Limitations}
\label{app:limits}
\noindent\textbf{Evaluator.}
The evaluator treats objects as rigid bodies of uniform density and tests stability only with short, gentle pushes. It therefore counts the valid contact of soft objects as collision (\cref{fig:failure_cases}a), and it cannot detect slow tipping, failure under larger forces, or whether a support is strong enough. Its thresholds and size ranges are fixed, so violation rates are best read as comparisons between methods, and unusual but valid furniture can fail the scale test.

\medskip
\noindent\textbf{Input estimates.}
The floor, masks, and support relations are estimated from the image and kept fixed. Errors in them pass directly into evaluation and repair, as the missing pillow and the floating chair show (\cref{fig:failure_cases}).

\medskip
\noindent\textbf{Repair.}
Repair only moves, rotates, and rescales objects, within small limits and one edit at a time. It cannot fix wrong shapes or object counts, cannot fully fix a badly wrong initial layout, and may stop when an error requires several objects to move together.

\medskip
\noindent\textbf{Cost and scope.}
The full pipeline takes about 15.07\% of PAT3D's reported per-scene runtime, although the timings come from different experimental setups. 
It remains too slow for interactive use, and PhyMix applies only to scenes made of separate objects.

\medskip
\noindent\textbf{Object quality.}
Object quality is limited by the underlying 3D generator,
which may produce inaccurate geometry or appearance inconsistent with the input image.

\medskip
\noindent\textbf{Future work.}
Soft-body simulation, input estimates with confidence, and repair that can regenerate objects or plan several edits together would address these limitations.

\clearpage
{\small
\bibliographystyle{corlabbrvnat}
\bibliography{paper}
}
\end{document}